\documentclass{article}

\usepackage{microtype}
\usepackage[pdftex]{graphicx}
\usepackage{booktabs} 
\usepackage{caption}
\usepackage{subcaption}
\usepackage{amsmath}
\usepackage{amsthm}
\usepackage{booktabs}       
\usepackage{multirow}
\usepackage{amssymb}
\usepackage{bm}
 \usepackage{hyperref}

\newcommand{\frtrain}{FR-Train}
\newtheorem{thm}{Theorem}
\newtheorem{definition}{Definition}
\usepackage{cancel}
\usepackage{color}
\usepackage{ulem,xpatch}
\xpatchcmd{\sout}
  {\bgroup}
  {\bgroup}
  {}{}

\newcommand{\E}{\mathbb{E}}
\newcommand{\KL}{D_\mathrm{KL}}

\usepackage[accepted]{icml2020}

\icmltitlerunning{\frtrain{}: A Mutual Information-Based Approach to Fair and Robust Training}

\begin{document}

\twocolumn[
\icmltitle{\frtrain{}: A Mutual Information-Based Approach to Fair and Robust Training}



\icmlsetsymbol{equal}{*}

\begin{icmlauthorlist}
\icmlauthor{Yuji Roh}{kaist}
\icmlauthor{Kangwook Lee}{wisconsin}
\icmlauthor{Steven Euijong Whang}{kaist}
\icmlauthor{Changho Suh}{kaist}
\end{icmlauthorlist}

\icmlaffiliation{kaist}{School of Electrical Engineering, Korea Advanced Institute of Science and Technology (KAIST), Daejeon, Korea}
\icmlaffiliation{wisconsin}{Department of Electrical and Computer Engineering, University of Wisconsin-Madison, Madison, Wisconsin, USA}

\icmlcorrespondingauthor{Steven Euijong Whang}{swhang@kaist.ac.kr}




\icmlkeywords{fairness, robustness, mutual information}

\vskip 0.3in
]



\printAffiliationsAndNotice{}  

\begin{abstract}
Trustworthy AI is a critical issue in machine learning where, in addition to training a model that is accurate, one must consider both {\em fair} and {\em robust} training in the presence of data bias and poisoning. 
However, the existing model fairness techniques mistakenly view poisoned data as an additional bias to be fixed, resulting in severe performance degradation. 
To address this problem, we propose \frtrain{}, which {\em holistically performs fair and robust model training}. 
We provide a mutual information-based interpretation of an existing adversarial training-based fairness-only method, and apply this idea to architect an additional discriminator that can identify poisoned data using a clean validation set and reduce its influence. 
In our experiments, \frtrain{} shows almost no decrease in fairness and accuracy in the presence of data poisoning by both mitigating the bias and defending against poisoning.
We also demonstrate how to construct clean validation sets using crowdsourcing, and release new benchmark datasets\footnote{\href{https://github.com/yuji-roh/fr-train}{https://github.com/yuji-roh/fr-train}}.
\end{abstract}

\section{Introduction}
\label{sec:introduction}
As machine learning becomes widespread in the Software 2.0 era~\cite{karpathi}, {\em trustworthy AI} is becoming increasingly critical. In addition to simply training accurate models, there is an urgent need to address multiple requirements including fairness, robustness, explainability, transparency, and accountability altogether~\cite{trustedai}. In particular, we focus on fairness and robustness, which are closely related issues that are affected by the same training data. For sensitive applications like healthcare, finance, and self-driving cars, a trained model must not discriminate customers based on sensitive attributes including age, sex, or religion. In addition, as applications often rely on external datasets for their training data, the model training must be resilient against noisy, subjective, or even adversarial data.

Traditionally, model fairness research~\cite{DBLP:conf/pods/Venkatasubramanian19,DBLP:journals/corr/abs-1810-08810,DBLP:conf/icse/VermaR18} has focused on developing metrics such as disparate impact~\cite{DBLP:conf/kdd/FeldmanFMSV15}, equalized odds~\cite{DBLP:conf/nips/HardtPNS16}, and equal opportunity~\cite{DBLP:conf/nips/HardtPNS16}, which capture various notions of discrimination. More recently, there has been a surge in {\em unfairness mitigation} techniques~\cite{DBLP:journals/corr/abs-1810-01943}, which improve the model fairness by either fixing the training data, training process, or trained model. Unfairness mitigation usually involves some tradeoff between the model's accuracy and fairness. Most recently, generative adversarial networks (GANs) are being adapted to a fairness setting~\cite{DBLP:conf/aies/ZhangLM18}. 
The architecture of GANs is suitable because accuracy and fairness are not always aligned, and it makes sense to simultaneously train two models: a classifier that predicts labels using input features and an adversary that predicts sensitive attributes using the classifier's predicted labels.

Robust model training is also important and needs to be concurrently taken into consideration. As dataset publishing is becoming mainstream as demonstrated by systems like Kaggle and Google Dataset Search~\cite{47845}, it is easy to publish data that is noisy, subjective, and even adversarial, which we hereafter refer to as {\em poisoned data}. As a result, there has been a proliferation of algorithms that make model training resilient to data poisoning as well~\cite{DBLP:conf/nips/NatarajanDRT13,DBLP:journals/jmlr/BiggioNL11,DBLP:journals/tnn/FrenayV14}. 
However, data poisoning attacks have become increasingly sophisticated, and defending against all of them is difficult~\cite{DBLP:journals/corr/abs-1811-00741}.

\begin{figure}[t]
\centering
\includegraphics[width=0.97\columnwidth]{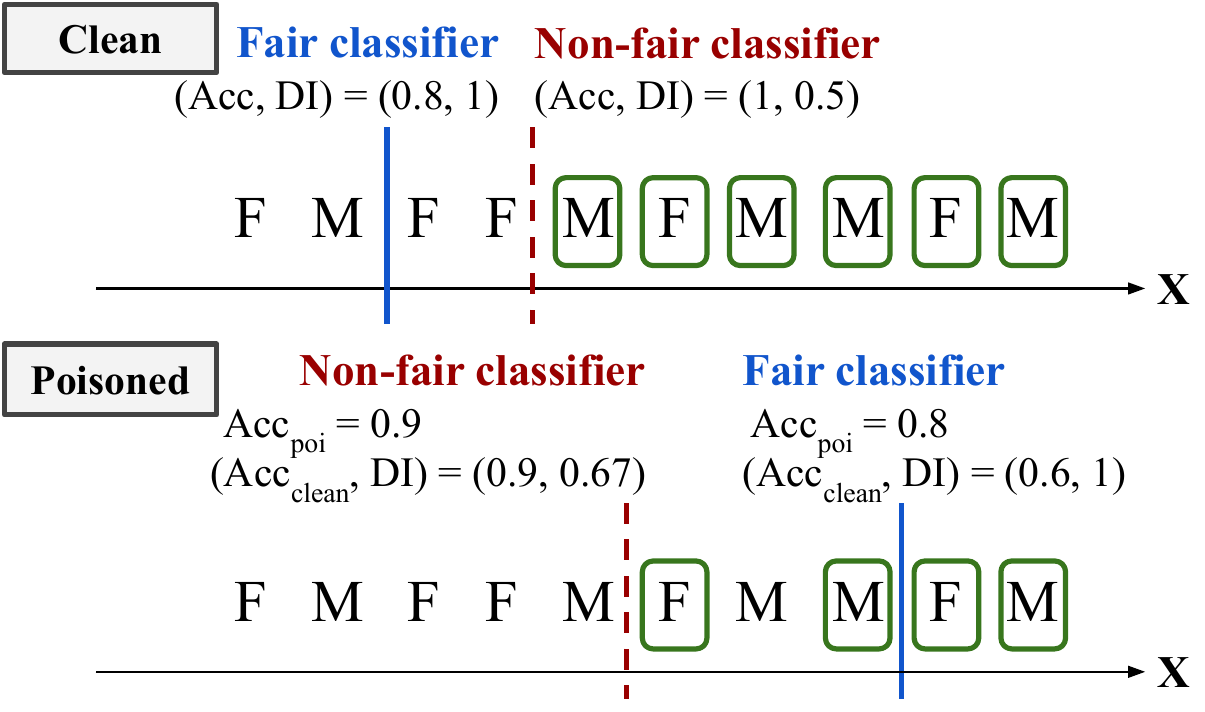}
\vspace{-0.4cm}
\caption{A small dataset of 10 people who need loans (F: female, M: male). A rounded box indicates a positive label. The clean data (above) is poisoned by flipping two labels (below). The vertical lines are the decision boundaries of non-fair and fair threshold classifiers. DI is disparate impact, and Acc$_{\text{clean}}$ (Acc$_{\text{poi}}$) is the accuracy on clean (poisoned) data.}
\vspace{-0.5cm}
\label{fig:toy_example}
\end{figure}

Solving model fairness without addressing data poisoning may lead to a worse tradeoff between accuracy and fairness. For example, consider a banking system that is giving out loans where there are two sensitive groups: men and women. Suppose we use disparate impact~\cite{DBLP:conf/kdd/FeldmanFMSV15} as the fairness measure. If the model's positive prediction rate is $M$ for men and $W$ for women, the disparate impact is $\min\{\frac{M}{W}, \frac{W}{M}\}$ where a value of 1 is considered perfectly fair. Figure~\ref{fig:toy_example} shows a toy example of five men and five women who need loans.
Each person is associated with a single-dimensional feature $x$, and only the ones with a rounded box would pay back their loans (i.e., their labels are positive). Let us train a threshold classifier that divides the people into two groups where those on the left are denied loans and those on the right are granted loans. On the clean data above, a classifier that does not consider fairness (non-fair classifier, red dotted line) can have perfect accuracy at the cost of having a disparate impact of 0.5 because 40\% of females are granted loans while 80\% of males are granted loans. On the other hand, a fair classifier (blue solid line) can divide the people such that the disparate impact is perfect, but the accuracy is only 0.8. Now suppose we poison the data where we flip the labels of the 5$^{th}$ and 7$^{th}$ persons (both male) from positive to negative as shown below. While each classifier is trained on the poisoned data, its accuracy is measured using the clean data labels. For the non-fair classifier trained on this data, the results are mixed where the accuracy decreases from 1 to 0.9, but the disparate impact increases from 0.5 to 0.67. However, the fair classifier has strictly worse results where the accuracy decreases from 0.8 to 0.6 without any change in the disparate impact. Hence, the fair classifier's accuracy-fairness tradeoff is worse when the data is poisoned.
One proposal is to sanitize the data prior to the model training, but it is known that removing poisoning without any knowledge of the model is extremely difficult~\cite{DBLP:journals/corr/abs-1811-00741}.

Our main contribution is an integrated solution called \frtrain{}, which trains accurate models that are also fair and robust to poisoning. \frtrain{} extends a state-of-the-art fairness-only method called Adversarial Debiasing (AD)~\cite{DBLP:conf/aies/ZhangLM18}, which consists of a generator used for classification and a discriminator that distinguishes predictions from one sensitive group against others, similar to GANs~\cite{DBLP:conf/nips/GoodfellowPMXWOCB14}. The discriminator ensures that the prediction $\hat{y}$ is independent of the sensitive attribute $z$. We first provide interpretation of such an adversarial learning approach using mutual information. We then use the results as an inspiration to add a new robustness discriminator that uses mutual information to distinguish (training examples, predictions) of the training data from (validation examples, validation labels) of a separate and clean validation set. This discriminator ensures that the model predictions on the training data are ``consistent'' with labels on clean data, where the clean validation set acts as a reference to the training. In addition, we also utilize the robustness discriminator results to further improve the fairness training by re-weighting examples. In our experiments, we show that addressing robustness and fairness sequentially during model training is not as effective as addressing them concurrently as in \frtrain{}.

Another contribution is addressing the challenge of constructing a clean validation set and gracefully handling the case where it is small or unavailable. To this end, we demonstrate a practical crowdsourcing method using majority voting for constructing a clean validation set, which has less poisoning than the input data. We construct clean validation sets from real datasets using Amazon Mechanical Turk and release them as a community resource. In the worst case when the validation set is non-existent, we show how the parameters of \frtrain{} can be adjusted to still maintain reasonable accuracy and fairness. 

In the following sections, we demonstrate the weaknesses of current fairness methods, propose \frtrain{} with experiments, and present the related work.

\section{Vulnerability of Fairness Methods}
\label{sec:vulnerability}

\begin{figure}[t]
\centering
\begin{subfigure}{\columnwidth}
\centering
\includegraphics[scale=0.23]{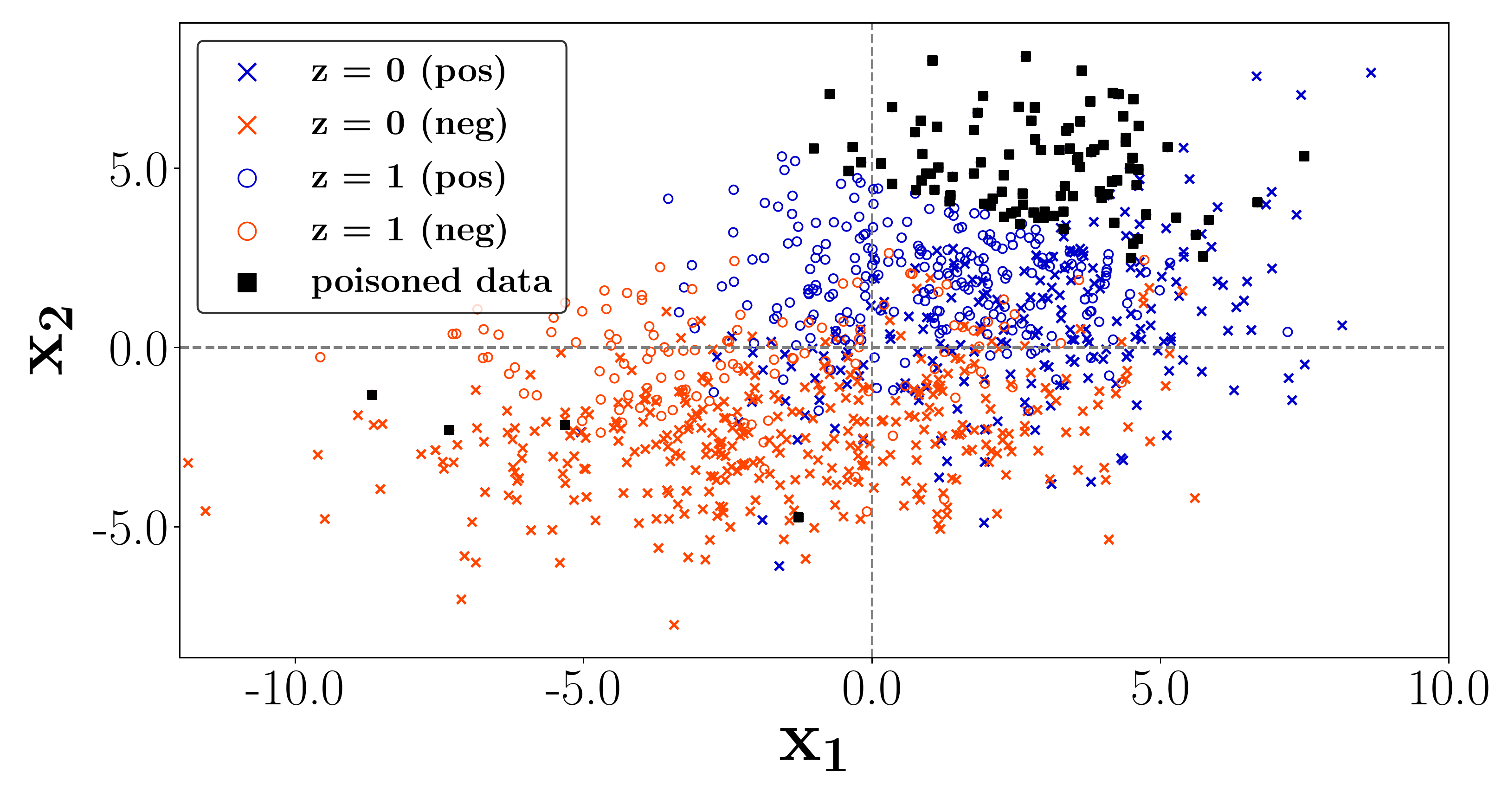}
\vspace{-0.2cm}
\caption{Synthetic data with label-flipped poisoning}
\label{fig:syntheticdatapoisoning}
\end{subfigure}
\begin{subfigure}{\columnwidth}
\centering
\includegraphics[scale=0.23]{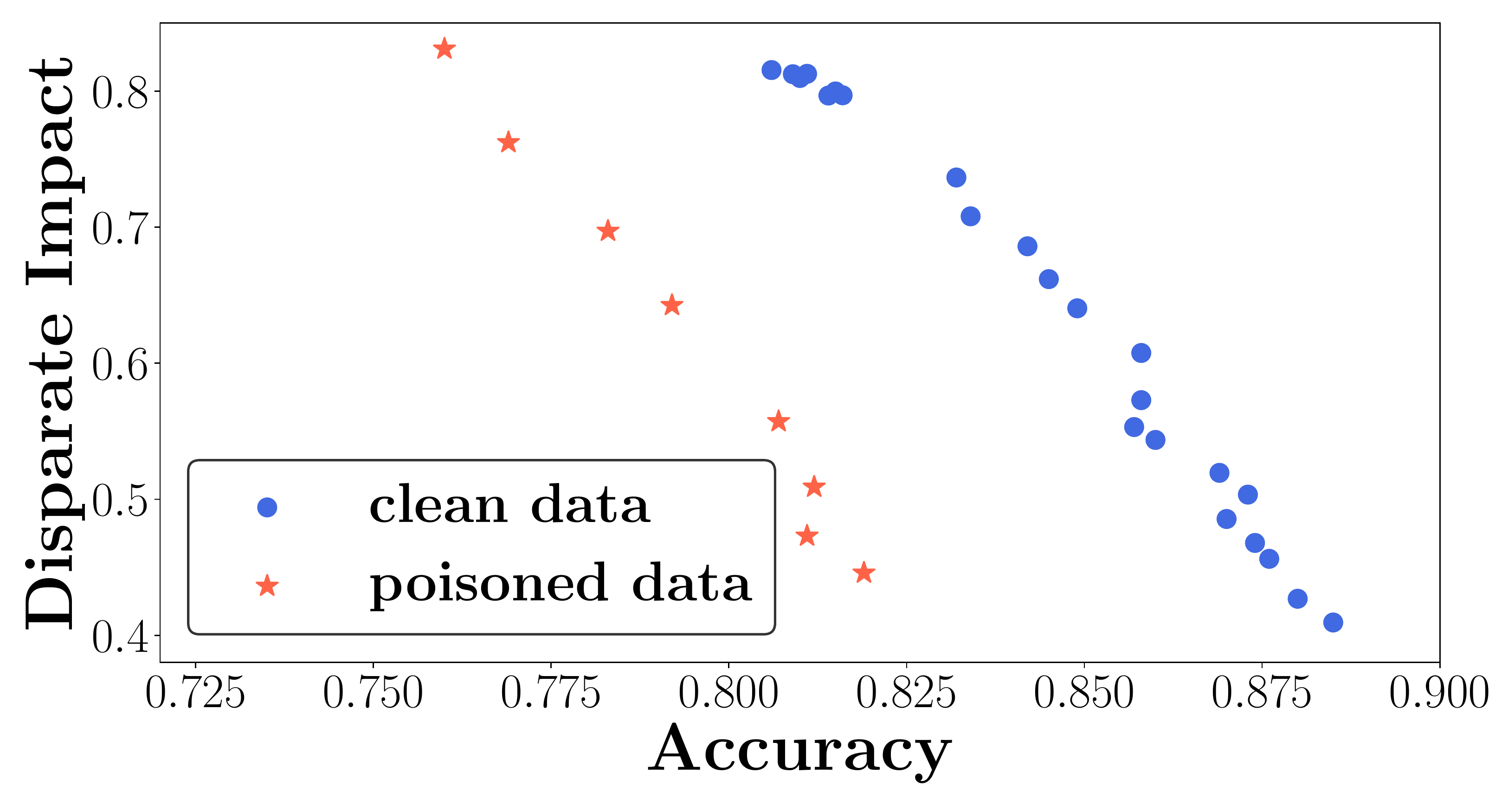}
\vspace{-0.2cm}
\caption{Accuracy-fairness tradeoff curves for Fairness Constraints}
\label{fig:fairnessconstraints}
\end{subfigure}
\caption{The top figure shows a synthetic dataset with data poisoning. Examples are divided into $z$ = 1 (marked with circles) and $z$ = 0 (crosses) as per a sensitive attribute $z$. The blue points indicate positive labels while the red points denote negative ones. For the poisoning, we flipped labels of $10\%$ of the examples with $z$ = 1 so as to maximize the accuracy performance degradation~\cite{DBLP:conf/pkdd/PaudiceML18}. The bottom figure shows that poisoning significantly worsens the accuracy-fairness tradeoff (i.e., the curve shifts to the left) of the Fairness Constraints method~\cite{DBLP:conf/aistats/ZafarVGG17}.}
\vspace{-0.5cm}
\end{figure}

We perform experiments to demonstrate that state-of-the-art fairness methods are indeed vulnerable even to simple poisoning attacks. We generate a synthetic dataset as shown in Figure~\ref{fig:syntheticdatapoisoning} (see the generation details in Section~\ref{sec:syntheticdataresults}). There are two non-sensitive attributes $x_1$ and $x_2$, which are reflected in the x-axis and y-axis, respectively. The examples are further divided into two classes based on the {\em sensitive} attribute $z$.  For generation of poisoned data, we poison 10\% of the training data by flipping the labels of examples that belong to a specific $z$ attribute (for this experiment $z$ = 1) so as to maximize the accuracy performance degradation. This approach is similar to an existing label flipping method~\cite{DBLP:conf/pkdd/PaudiceML18}. To make a validation set, we randomly select clean examples that amount to 10\% of the entire training data.

We use disparate impact as the fairness measure and evaluate a fairness method called Fairness Constraints~\cite{DBLP:conf/aistats/ZafarVGG17}, which incorporates a regularization term that reflects fairness constraints in the context of convex margin-based classifiers such as logistic regression and support vector machines (SVMs). As this method involves a  regularization factor $\lambda$ that balances the accuracy and fairness objectives, we can obtain a tradeoff curve by adjusting its value.
Figure~\ref{fig:fairnessconstraints} shows two accuracy-fairness tradeoff curves obtained with the clean and poisoned synthetic datasets. 
Notice that adding data poisoning clearly shifts the curve to the left, which means accuracy decreases. 
This coincides with our intuition. The poisoning confuses the model so that there are more biased examples to fix, which in turn makes it {\em overreact} and thus sacrifice more on accuracy. We also leave in the supplementary the accuracy-fairness tradeoff curves of Fairness Constraints on real datasets. The results clearly show that both accuracy and fairness decrease on the poisoned data. In Section~\ref{sec:experiments}, we will show how data poisoning affects other fairness methods.

\section{\frtrain{}}

\begin{figure}[t]
  \centering
     \includegraphics[width=\columnwidth]{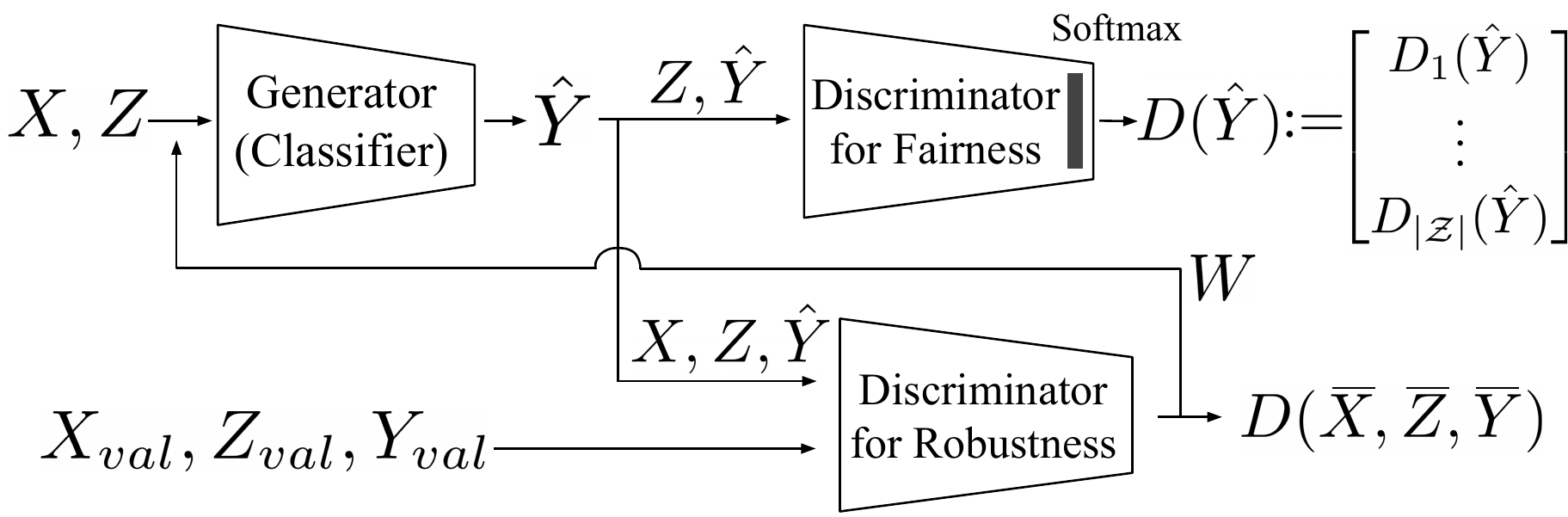}
      \vspace{-0.4cm}
     \caption{The architecture of \frtrain{}.}
 \label{fig:frgan}
 \vspace{-0.4cm}
\end{figure}

We now describe \frtrain{} (see Figure~\ref{fig:frgan}). Unlike traditional GANs, the generator is a classifier that receives an example $x \in X$ and returns a prediction $\hat{y}$. There are two discriminators that respectively optimize fairness and robustness using mutual information. In addition, the outputs of the robustness discriminator can be used to further improve the fairness training by re-weighting examples.

\subsection{Fairness}
\label{sec:fairness}

We denote by $\mathcal{D}_{tr}$ the training data set.
Suppose $\mathcal{D}_{tr}$ has $m$ examples \{($x^{(i)}$, $z^{(i)}$,  $y^{(i)}$)\}$_{i=1}^m$ where $x^{(i)}$ contains the non-sensitive attributes, $z^{(i)}$ contains the sensitive attributes, and $y^{(i)}$ is the label. Both the sensitive attribute and label can be multi-class, i.e., they can have one of multiple values. For notational simplicity, we assume there is one sensitive attribute, which can be viewed as a merged result of multiple sensitive attributes with a larger alphabet size. For illustrative purposes, we focus on disparate impact, leaving in the supplementary our formulation and experimental results for equalized odds and equal opportunity. 
Disparate impact aims for the same positive prediction ratio for each sensitive attribute $z \in \mathcal{Z}$ where $\mathcal{Z}$ is the set of possible sensitive attribute values. We use the following definition for disparate impact:

\begin{definition}{(Disparate Impact)}
\\$P(\hat{Y}=1|Z=z_1) = P(\hat{Y}=1|Z=z_2),~\forall z_1, z_2 \in \mathcal{Z}$. 
\end{definition}

The first discriminator in \frtrain{} distinguishes predictions w.r.t.\@ one sensitive group from those in the others. Disparate impact intends the sensitive attribute to be independent of the model's prediction, i.e., $I(Z;\hat{Y}) = 0$.

We explain how \frtrain{} can enforce the above constraint. Let $P_Z(z)$ be the distribution of $Z$ where $z \in \mathcal{Z}$. Let $\hat{Y} | Z = z \sim P_{\hat{Y}|z}(\cdot)$ and $\hat{Y} \sim P_{\hat{Y}}(\cdot)$. Then $P_{\hat{Y}}(\cdot) = \sum_{z \in \mathcal{Z}} P_Z(z) P_{\hat{Y}|z}(\cdot)$.

The following theorem asserts that mutual information is equivalent to the following function optimization where the optimal discriminator $D_z^{\star}(\hat{y}) = P_{Z|\hat{Y}} ( z|\hat{y} )$ and $\sum_{z \in \mathcal{Z}} D_z^{\star}(\hat{y}) = 1,~\forall \hat{y} \in \mathcal{Y}$. 
\begin{thm}
\label{thm:mi}
$I(Z;\hat{Y}) =$\\ 
\resizebox{\hsize}{!}{$\max \limits_{D_z(\hat{y}):\sum_z D_z(\hat{y})=1,~\forall \hat{y}} \sum \limits_{z \in \mathcal{Z}} P_Z(z) \mathbb{E}_{P_{\hat{Y}|z}} \left[ \log D_z(\hat{Y}) \right]+H(Z).$}
\end{thm}
While deferring the detailed proof to the supplemental materials, we provide a brief overview of the proof.
As the optimization problem in the RHS is convex, we find the optimal discriminator by solving the KKT conditions. 
We then show that the maximum value attained by the optimal discriminator is equal to the mutual information by using the properties of mutual information and the generalized Jensen-Shannon divergence~\cite{lin1991divergence}. 

What is more involved than showing the above equality is designing the right optimization problem. 
One needs to carefully handcraft a plausible optimization problem so that its unique solution matches the desired quantity. 
Here, we design the optimization problem via a `guess-\&-check' approach aided by the structural insights across the KL divergences that appear in an alternative expression of mutual information.

We now discuss how to implement the above expression. Since we do not know $P_{\hat{Y}|z}(\cdot)$ exactly, we compute the following empirical version:
\begin{align*}
    \resizebox{\hsize}{!}{$\max\limits_{D_z(\hat{y}):\sum_z D_z(\hat{y})=1,~\forall \hat{y}} \sum\limits_{z \in \mathcal{Z}} P_Z(z) \sum\limits_{i:z^{(i)}=z} \frac{1}{m_z} \log D_z(\hat{y}^{(i)}) + H(Z)$}.
\end{align*}
Now for sufficiently large $m$, the number $m_z$ of examples with $z^{(i)} = z$ is approximately the same as $P_Z(z)m$. Therefore, the above expression becomes:
\begin{align*}
    \max\limits_{D_z(\hat{y}):\sum_z D_z(\hat{y})=1,~\forall \hat{y}}\sum\limits_{z \in \mathcal{Z}} \sum\limits_{i:z^{(i)}=z} \frac{1}{m} \log D_z(\hat{y}^{(i)}) + H(Z).
\end{align*}
Interestingly, this formulation is exactly the same as that in the original GAN~\cite{DBLP:conf/nips/GoodfellowPMXWOCB14} when $|\mathcal{Z}| = 2$. We also remark that our formulation does not require a prior knowledge on $P_Z(z)$.

We note that Adversarial Debiasing (AD)~\cite{DBLP:conf/aies/ZhangLM18} has an additional projection term that is used to force the classifier to never decrease the discriminator's loss. However, we do not use this term in \frtrain{} because it worsens the training stability in our experiments. 

\subsection{Robustness}
\label{sec:robustness}
The robustness discriminator ensures robust training by using mutual information to distinguish examples and predictions from a clean validation set. For now, let us assume such a validation set exists (in Section~\ref{sec:realdata}, we demonstrate how to construct one). The discriminator then distinguishes the training data with predictions $\{(x^{(i)}, z^{(i)}, \hat{y}^{(i)}$)\}$_{i=1}^m$ from the validation set $\{(x_{\text{val}}^{(i)}, z_{\text{val}}^{(i)}, y_{\text{val}}^{(i)})\}_{i=1}^{m_{\text{val}}}$. 
Intuitively, if the classifier is confused by data poisoning in the training data, then its predictions will not be consistent with the labels of the clean data, and the discriminator would be able to detect that difference. 
Our use of a validation set is inspired by meta learning-based robust training algorithms~\cite{DBLP:conf/icml/RenZYU18}, which also defends against poisoning attacks by using the validation data loss as a meta objective. However, a key difference is that we take an adversarial learning approach, which introduces a knob that controls the emphasis of robust training. We find that this knob enables \frtrain{} to be more robust to the validation set size (see details in Section~\ref{sec:syntheticdataresults}). In Section~\ref{sec:architecture}, we also use the robustness discriminator to further improve the fairness training using example re-weighting.

We first define $\overline{X} = VX+(1-V)X_\text{val}, \overline{Z} = VZ+(1-V)Z_\text{val}$, and $\overline{Y} = V\hat{Y}+(1-V)Y_\text{val}$.
Here, note that $V$ is an indicator random variable that denotes whether an example is generated ($V = 1$) or comes from the validation set ($V = 0$). 
We then want to ensure that the distribution of $(X, Z, \hat{Y})$ matches that of $(X_\text{val}, Z_\text{val}, Y_\text{val})$.
This can be done by enforcing $I(V; \overline{X}, \overline{Z}, \overline{Y}) = 0$, i.e., the predictions on the training data are indistinguishable from the labels of the validation set. 
Thus we can mimic the clean dataset while expecting an indirect sanitization effect.

Analogous to the fairness discriminator, we show that mutual information is equivalent to the following function optimization where the optimal discriminator $D_v^{\star}(x,z,y) = P_{V|\overline{X}, \overline{Z}, \overline{Y}} (v|x,z,y)$ and $\sum_{v \in \mathcal{V}} D_v^{\star}(x,z,y) = 1,~\forall (x,z,y) \in \mathcal{X} \times \mathcal{Z} \times \mathcal{Y}$. The proof is similar to that of Theorem~\ref{thm:mi}. 

\begin{thm}
\label{thm:mi2}
$I(V; \overline{X}, \overline{Z}, \overline{Y})$ =\\
$\max \limits_{D_v (x, z, y): \sum_v D_v(x, z, y) = 1,~\forall (x,z,y)}$\\
$\sum\limits_{v \in \mathcal{V}}P_V(v)\mathbb{E}_{P_{\overline{X}, \overline{Z}, \overline{Y}|v}} \left[ \log D_v(\overline{X}, \overline{Z}, \overline{Y}) \right] + H(V).$
\end{thm}

\subsection{Architecture}
\label{sec:architecture}

We describe the \frtrain{} architecture in Figure~\ref{fig:frgan}. For the loss function of the generator, we employ cross entropy:
\begin{align*}
    L_1 = \frac{1}{m} \sum_{i=1}^m -y^{(i)} \log \hat{y}^{(i)} - (1 - y^{(i)})\log (1 - \hat{y}^{(i)}).
\end{align*}
We set the loss function w.r.t.\@ the fairness discriminator as:
\begin{align*}
L_2 = \max_{D(\cdot)} \sum_{z \in \mathcal{Z}} \sum_{i:z^{(i)}=z} \frac{1}{m} \log D_z(\hat{y}^{(i)}) + H(Z)
\end{align*}
where $D(\cdot):=(D_1(\cdot), \dots, D_{|\cal Z|}(\cdot))$. The condition $\sum_{z \in \mathcal{Z}} D_z^{\star}(\hat{Y}) = 1$ can be enforced by adding a softmax layer to the discriminator.

Finally, implementing $I(V; \overline{X}, \overline{Z}, \overline{Y})$, we set the loss function w.r.t. the robustness discriminator as:
\begin{align*}
L_3 = & \max_{D^r(\cdot)} \sum_{i:v^{(i)} = 0} \frac{1}{m}\log D^r(x^{(i)}_{val},z^{(i)}_{val},y^{(i)}_{val}) +  \\
 & \sum_{i:v^{(i)}=1} \frac{1}{m} \log(1 - D^r(x^{(i)},z^{(i)},\hat{y}^{(i)})) + H(V).
\end{align*}
The final objective function is the weighted sum of these value functions:
\begin{align*}
    \min_{G(\cdot)}L_1 + \lambda_1 L_2 + \lambda_2 L_3.
\end{align*}
Here $\lambda_1$ and $\lambda_2$ are tuning knobs that play roles to emphasize fair and robust training, respectively.

\paragraph{Example Re-weighting for Fairness Training}

In addition to the above architecture, we also utilize the decision values $D^r(X, Z, \hat{Y})$ of the robustness discriminator as example weights to further improve the fairness training (in Figure~\ref{fig:frgan}, the arrow from the robustness discriminator's output to the classifier's input). In particular, the two losses $L_1$ and $L_2$ are now computed using the example weights. The intuition is that, by giving more weight to the clean examples, we can improve the accuracy-fairness tradeoff. A question is when to apply these weights. If we apply the weights too early, then $D(X, Z, \hat{Y})$ may not be accurate enough and actually harm the fairness training. Intuitively, we would like to use the discriminator's results when we know it is performing at least as well as the classifier. Hence, for a more reliable signal, we use the {\em relative performance} between the classifier and robustness discriminator to generate the weights. Given the classifier's loss $L_c$ and the robustness discriminator's loss $L_d$, we compute the final example weights as $W = R + D(X, Z, \hat{Y}) \times (1 - R)$ where $R = \sigma(\frac{L_c}{L_d} - C)$ is a conversion of the loss ratio into a probability using the sigmoid function $\sigma$ and hyperparameter~$C$.
We note that $C$ acts as a threshold on the loss ratio.

\section{Experiments}
\label{sec:experiments}

We provide experimental results for \frtrain{}. For the fairness measure, we use disparate impact, while leaving in the supplementary the results for equalized odds and equal opportunity. We evaluate all models on separate clean test sets.
In our experiments, we use two sensitive attributes $z_1$ and $z_2$, and disparate impact is measured as the ratio $\min \{ \frac{P(\hat{Y}=1|Z=z_1)}{P(\hat{Y}=1|Z=z_2)}, \frac{P(\hat{Y}=1|Z=z_2)}{P(\hat{Y}=1|Z=z_1)} \}$.
We use PyTorch~\cite{paszke2017automatic}, and all experiments are performed on a server with Intel i7-6850 CPUs. More implementation details are in the supplementary.

\subsection{Synthetic Data Results}
\label{sec:syntheticdataresults}

For the synthetic data, we generate 2,000 examples with two non-sensitive attributes $x_1$ and $x_2$, a sensitive attribute $z$, and a label $y$, using a method similar to the algorithm proposed by~\cite{DBLP:conf/aistats/ZafarVGG17}. 
Both $z$ and $y$ are binary, and the ($x_1$, $x_2$) pair consists of two normal distributions: $(x_1,x_2)|y=0 \sim \mathcal{N}([-2; -2], [10, 1; 1, 3])$ and $(x_1,x_2)|y=1 \sim \mathcal{N}([2; 2], [5, 1; 1, 5])$. The $z$ attribute has the Bernoulli distribution $p(z=1)=p((x'_1, x'_2)|y=1)/[p((x'_1, x'_2)|y=0)+p((x'_1, x'_2)|y=1)]$ where $(x'_1, x'_2)=(x_1\cos(\pi/4)-x_2\sin(\pi/4), x_1\sin(\pi/4)+x_2\cos(\pi/4))$.
Finally for each example, the $x_1$ and $x_2$ values are sampled as per the normal distribution associated with the $y$. For data poisoning, we flip the labels of examples with $z = 1$ so as to maximize the accuracy performance degradation as described in Section~\ref{sec:vulnerability}, and the amount of poisoning is 10\% of $\mathcal{D}_{tr}$. In the supplementary, we also perform \frtrain{} varying the amount of poisoning from 10\% to 40\%.

\begin{table}[t]
  \caption{Accuracy and fairness performances on the synthetic test datasets w.r.t.\@ disparate impact (DI). Two types of methods are compared: (1) fairness methods: FC~\cite{DBLP:conf/aistats/ZafarVGG17}, LBC~\cite{DBLP:journals/corr/abs-1901-04966}, and AD~\cite{DBLP:conf/aies/ZhangLM18} where ``{\sc RML+}'' denotes the application of sanitization using RML~\cite{DBLP:conf/icml/RenZYU18} beforehand; (2) non-fairness methods: LR and RML. For \frtrain{} and RML, the validation set is 10\% of $\mathcal{D}_{tr}$. The amount of poisoning is 10\% of $\mathcal{D}_{tr}$. For each result of the poisoned data, we make a comparison with the clean data result and show the percentage increase or decrease.}
  \label{tbl:fairnesstradeoff}
  \centering
  \begin{tabular}{l@{\hspace{7pt}}c@{\hspace{7pt}}c@{\hspace{12pt}}c@{\hspace{7pt}}c}
    \toprule
     Method & \multicolumn{2}{c}{Clean data} & \multicolumn{2}{c}{Poisoned data} \\
    \cmidrule(r){1-5}
       & DI & Acc. & DI & Acc. \\
    \midrule
    FC & .822 & .806 & .831 (1.1\% ${\color{blue}\uparrow}$) & .760 (5.7\% ${\color{red}\downarrow}$) \\
    LBC & .819 & .760 & .827 (1.0\% ${\color{blue}\uparrow}$) & .715 (5.9\% ${\color{red}\downarrow}$)   \\
    AD & .807 & .811 & .834 (3.4\% ${\color{blue}\uparrow}$) & .769 (5.2\% ${\color{red}\downarrow}$)     \\
    RML+FC & .822 & .806 & .802 (2.4\% ${\color{red}\downarrow}$) & .529 (34.\% ${\color{red}\downarrow}$)    \\
    RML+LBC & .819 & .760 & .810 (1.1\% $\color{red}\downarrow$) & .752 (1.1\% $\color{red}\downarrow$)     \\
    RML+AD & .807 & .811 & .808 (0.1\% ${\color{blue}\uparrow}$) & .756 (6.8\% ${\color{red}\downarrow}$) \\
    \cmidrule(l){1-5}
    LR & .409 & .885 & .446 (9.1\% ${\color{blue}\uparrow}$) & .819 (7.5\% ${\color{red}\downarrow}$)  \\
    RML  & .471 & .876 & .395 (16.\% ${\color{red}\downarrow}$) & .869 (0.8\% ${\color{red}\downarrow}$)  \\
    \cmidrule(l){1-5}
    \textbf{\frtrain{}} & \textbf{.818} & \textbf{.807} & \textbf{.827 (1.1\% ${\color{blue}\uparrow}$)} & \textbf{.814 (0.9\% ${\color{blue}\uparrow}$)}  \\ 

    \bottomrule
  \end{tabular}
  \vspace{-0.4cm}
\end{table}

\begin{figure*}[t]
\centering
\captionsetup[subfigure]{justification=centering}
\begin{subfigure}{0.185\textwidth}
\hspace{-1cm}
\includegraphics[scale=0.15]{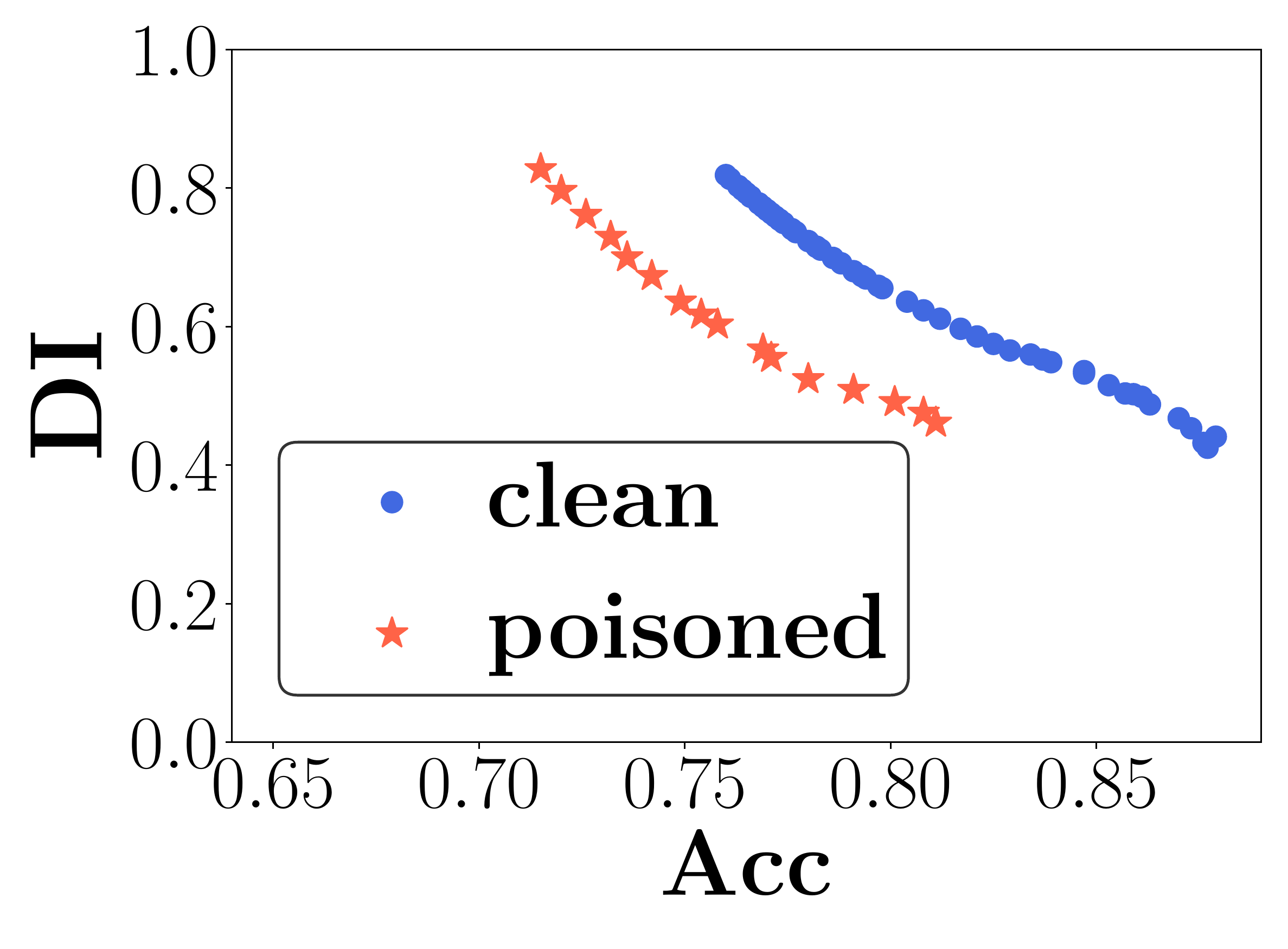}\vspace{-0.1cm}
\caption{LBC \newline}
\label{fig:labelbiascorrection}
\end{subfigure}
\begin{subfigure}{0.185\textwidth}
\hspace{-0.65cm}
\includegraphics[scale=0.15]{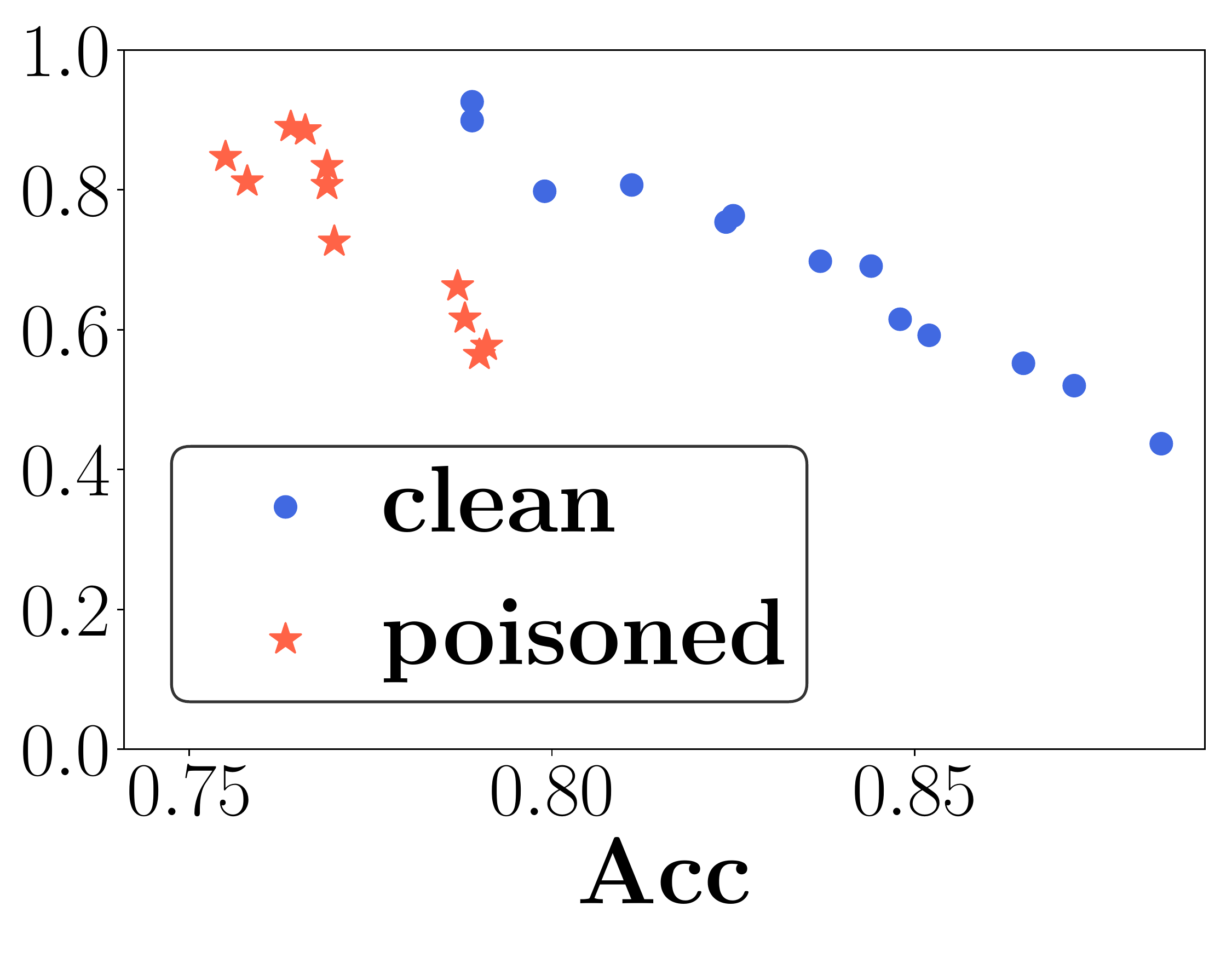}\vspace{-0.1cm}
\caption{AD \newline}
\label{fig:adversarialdebiasing}
\end{subfigure}
\begin{subfigure}{0.185\textwidth}
\hspace{-0.55cm}
\includegraphics[scale=0.15]{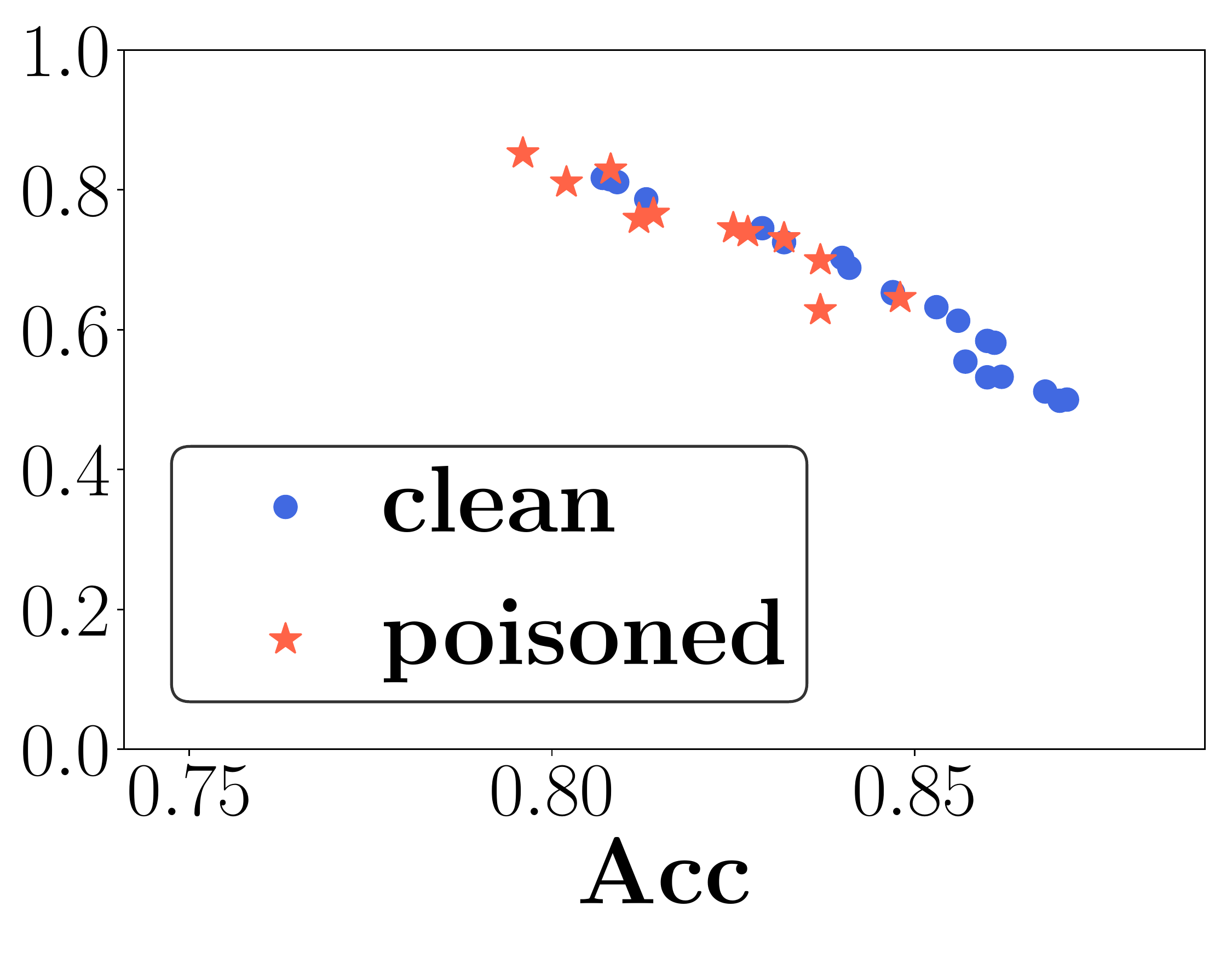}\vspace{-0.1cm}
\caption{\frtrain{} \\(Val. set size = 10\%)}
\label{fig:frtrain10}
\end{subfigure}
\begin{subfigure}{0.185\textwidth}
\hspace{-0.325cm}
\includegraphics[scale=0.15]{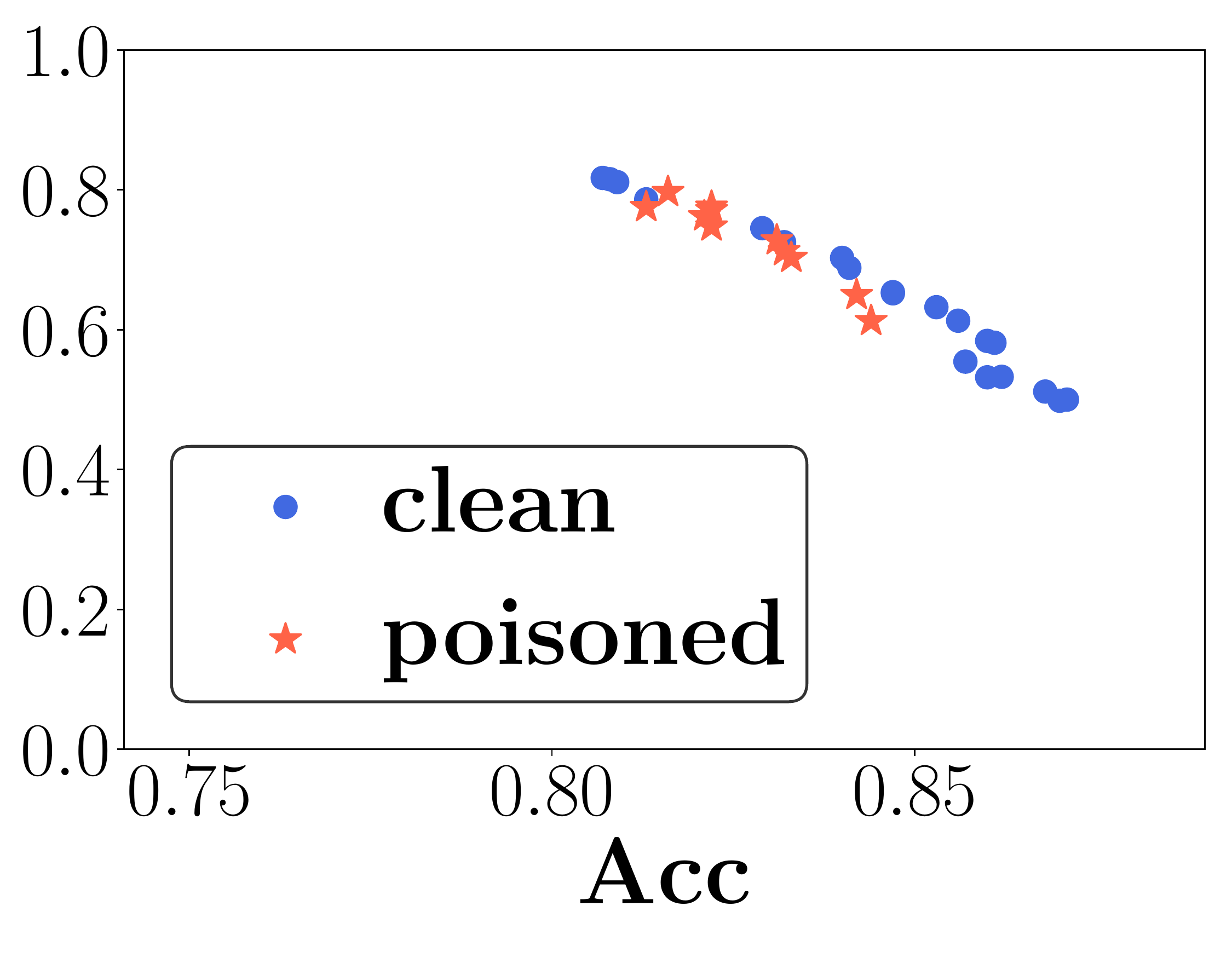}\vspace{-0.1cm}
\caption{\frtrain{} \\(Val. set size = 5\%)}
\label{fig:frtrain05}
\end{subfigure}
\begin{subfigure}{0.185\textwidth}
\includegraphics[scale=0.15]{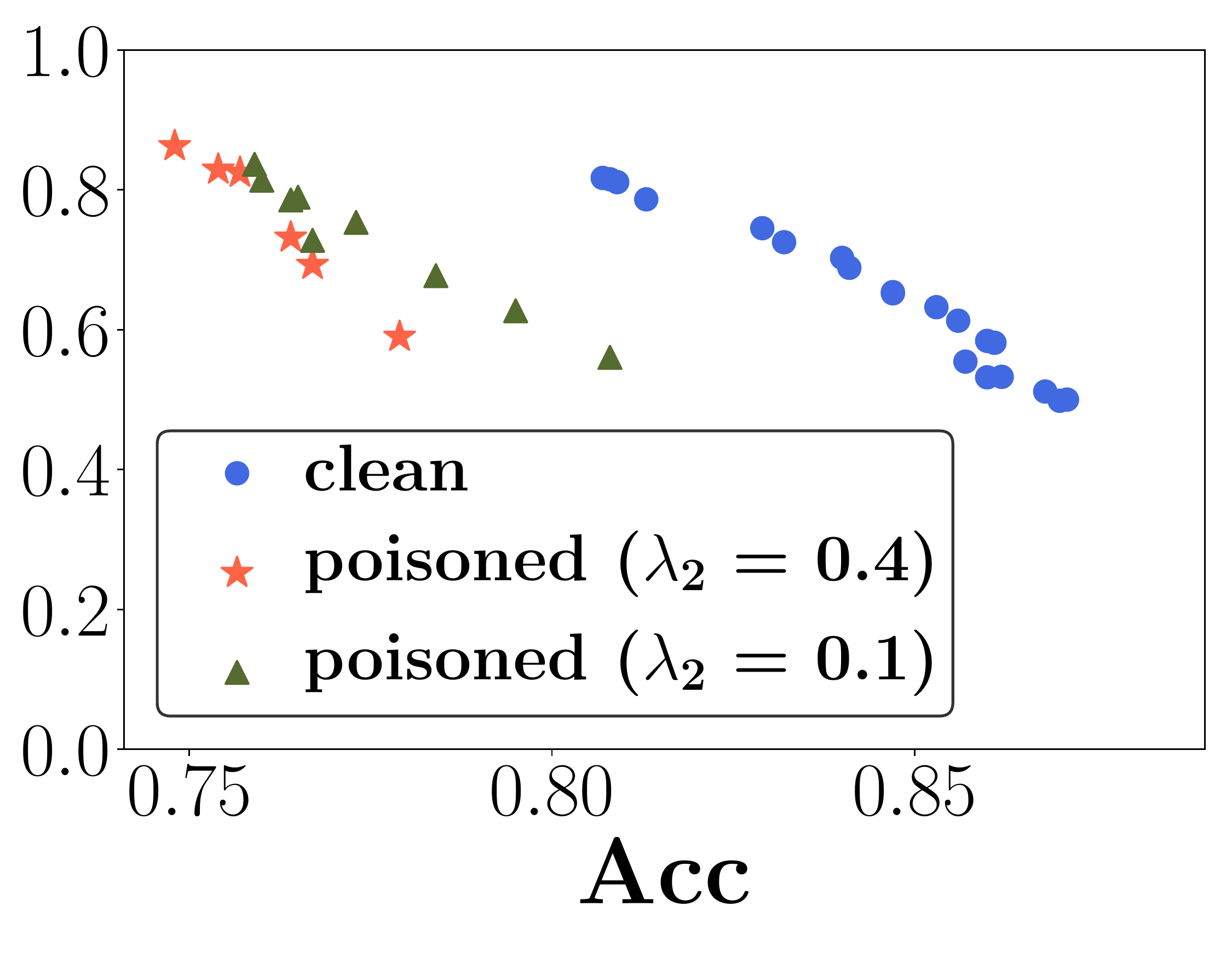}\vspace{-0.1cm}
\caption{\frtrain{} \\(Val. set size = 0.1\%)}
\label{fig:frtrain01}
\end{subfigure}

\label{fig:tradeoffs}
\vspace{-0.2cm}
\caption{Accuracy-fairness tradeoff curves. Figures (a) and (b) show that the poisoning worsens the accuracy-fairness tradeoffs of LBC~\cite{DBLP:journals/corr/abs-1901-04966} and AD~\cite{DBLP:conf/aies/ZhangLM18}. Figures (c) and (d) show that \frtrain{} maintains the tradeoffs even with a 5\% validation set. When the validation set is too small (Figure (e)), \frtrain{} can adjust $\lambda_2$ to reduce the adverse effect on training.}
\vspace{-0.2cm}
\end{figure*}

\paragraph{Accuracy and Fairness}

We compare \frtrain{} with various baselines. First, there are the fairness methods: Fairness Constraints~\cite{DBLP:conf/aistats/ZafarVGG17} (FC), Label Bias Correction~\cite{DBLP:journals/corr/abs-1901-04966} (LBC), and Adversarial Debiasing~\cite{DBLP:conf/aies/ZhangLM18} (AD).
As described in the previous sections, FC adds a penalty term that captures the prediction differences across sensitive groups, while AD utilizes adversarial learning to achieve high fairness. LBC is an example re-weighting algorithm, which assumes the existence of true \emph{unbiased yet unknown} labels.
LBC provides theoretical guarantees that training on the resulting loss corresponds to training on the true unbiased labels, which yields a fair model.
While there exist other re-weighting techniques including~\cite{DBLP:conf/icml/AgarwalBD0W18}, we choose LBC because it performs the best in experiments~\cite{DBLP:journals/corr/abs-1901-04966}.

Since \frtrain{} is to our knowledge the first method to address both fairness and robustness in model training, there is no fairness method that also performs data sanitization using a clean validation set. However, \cite{DBLP:conf/icml/RenZYU18} is a state-of-the-art robust training method based on meta learning using a clean validation set, which we call RML. For a fair comparison, we thus extend the three fairness methods by first performing RML and then utilizing the example weights in the fairness training in a straightforward fashion. In addition, we compare with non-fairness methods: logistic regression (LR) and RML. 

Table~\ref{tbl:fairnesstradeoff} compares \frtrain{} with the baselines. We use a validation set that amounts to 10\% of $\mathcal{D}_{tr}$. We also apply proper hyperparameters so that the disparate impacts are similar (around $0.8$) across all methods, if possible.
When setting $\lambda_1$ and $\lambda_2$ for \frtrain{}, we usually fix $\lambda_2$ to some value and then adjust $\lambda_1$ using one-round cross validation. 
There is no hyperparameter tuning for logistic regression and the meta learning-based robust training algorithm, as they have no knobs for adjusting fairness. The results show that for the fairness methods, data poisoning aggravates accuracy-fairness trade-offs. For example, the accuracy of FC falls by 5.7\%, while the disparate impact of it remains a similar value. 
On the other hand, the performance for \frtrain{} does not degrade: disparate impact and accuracy increase by 1.1\% and 0.9\%, respectively. Table~\ref{tbl:fairnesstradeoff} also shows that combining the fairness methods with RML (rows 4--6) does not always yield better accuracy and fairness. In fact, using sanitization may lower the accuracy or fairness (e.g., RML+FC has an accuracy of 0.529 on poisoned data while FC has 0.760). The results suggest that removing poisoning and then bias is not that effective.

We observe how accuracy trades off with fairness on clean and poisoned datasets. The results for FC are shown in Figure~\ref{fig:fairnessconstraints}. 
For LBC, we employ the number of training as a knob to trade accuracy off fairness since LBC gradually improves fairness by repeatedly updating example weights per training.
As shown in Figure~\ref{fig:labelbiascorrection}, the tradeoff curve shifts to the left, which demonstrates a clear tradeoff degradation. For AD, we employ the $\alpha$ parameter~\cite{DBLP:conf/aies/ZhangLM18} analogous to $\lambda_1$ as a knob to trade accuracy off fairness. Figure~\ref{fig:adversarialdebiasing} shows the tradeoff curve again shifts to the left.

\paragraph{Validation Set Size}

Figures~\ref{fig:frtrain10} to \ref{fig:frtrain01} show how the validation set size affects the robustness of \frtrain{}.
In particular, we compare the accuracy-fairness tradeoff of \frtrain{} on clean data and that on poisoned data while varying the size of the validation set.
When running on poisoned data, we fixed $\lambda_2 = 0.4$ and varied $\lambda_1$. 
We see that even a 5\% validation set (Figure~\ref{fig:frtrain05}) is sufficient to maintain the accuracy and fairness obtained on the clean data. When using 0.1\% (Figure~\ref{fig:frtrain01}), the validation set is too small and has an adverse effect on the training. 
However, by decreasing the tuning knob $\lambda_2$ down to 0.1, we can de-emphasize robust training, thereby avoiding the adverse effect (Figure~\ref{fig:frtrain01}, green triangles). 
This is in contrast to RML, which suffers from a non-negligible performance degradation for a very small validation set. See details in the supplementary.

\subsection{Real Data Results}
\label{sec:realdata}
We use two real datasets: ProPublica COMPAS~\cite{Compas} and AdultCensus~\cite{DBLP:conf/kdd/Kohavi96}, which have 7,214 and 45,222 examples, respectively. We use the same preprocessing as in IBM's AI Fairness 360~\cite{aif360-oct-2018} and use the sensitive attribute {\sc SEX} for both datasets. 
For data poisoning, we use the same method employed on synthetic data: flipping the labels with $z = 1$ so as to maximize the accuracy performance degradation. 
The amount of poisoning is 10\% of $\mathcal{D}_{tr}$.

While we assumed that a small yet clean validation set is available in the previous synthetic data experiments, such an assumption does not hold in practice. 
Thus, for real-data experiments, we consider a scenario where one first constructs a small (which amounts to 5\% of $\mathcal{D}_{tr}$) validation set based on crowdsourcing, and then uses it for \frtrain{}.
We provide details on how to construct this validation set in Section~\ref{sec:crowd}.

Summarized in Tables~\ref{tbl:realCompas} and \ref{tbl:realAdult} are the fairness and accuracy performances of various training algorithms on the COMPAS and AdultCensus datasets, respectively. 
As in Table~\ref{tbl:fairnesstradeoff}, we apply proper hyperparameters so that the disparate impacts are similar across all distinct methods, both for the clean and poisoned datasets. The results are similar to Table~\ref{tbl:fairnesstradeoff}: the three fairness methods have worse disparate impact and accuracy due to data poisoning; LR and RML exhibit poor disparate impacts; and \frtrain{} again shows little degradation both in fairness and accuracy. Tables~\ref{tbl:realCompas} and \ref{tbl:realAdult} also show that combining the fairness methods with sanitization using RML (rows 4--6) does not always yield better accuracy and fairness and may even lower them, which is consistent with the results on synthetic data. 
One may wonder if the fairness baselines would perform better if they are trained on the clean validation set. In the supplementary, we show that the performances are actually worse than those in Tables~\ref{tbl:realCompas} and~\ref{tbl:realAdult}. This is because the clean validation set is too small to be used as a stand-alone train data. Indeed, a similar observation is made in~\cite{Zhang2018TrainingSD}.

\begin{table}[t]
 \caption{Accuracy and fairness performances on COMPAS test data w.r.t.\@ disparate impact (DI) where the training data is poisoned using the label flipping attack. Two types of methods are compared: (1) fairness methods: FC, LBC, and AD where ``{\sc RML+}'' denotes the application of sanitization using RML beforehand; (2) non-fairness methods: LR and RML. For \frtrain{} and RML, the validation set is 5\% of $\mathcal{D}_{tr}$. The amount of poisoning is 10\% of $\mathcal{D}_{tr}$. For each result of the poisoned data, we compare with the clean data result and show the percentage increase or decrease. }
  \label{tbl:realCompas}
  \centering
  \begin{tabular}{l@{\hspace{7pt}}c@{\hspace{9pt}}c@{\hspace{12pt}}c@{\hspace{7pt}}c}
    \toprule
     Method & \multicolumn{2}{c}{Clean data} & \multicolumn{2}{c}{Poisoned data}                    \\
    \cmidrule(r){1-5}
      & DI & Acc. & DI & Acc. \\
    \midrule
     FC & .777 & .682 & .794 (2.2\% ${\color{blue}\uparrow}$) & .612 (10.\% ${\color{red}\downarrow}$) \\
     LBC & .866 & .671 & .838 (2.8\% ${\color{red}\downarrow}$)& .671 (0.0\% -)  \\
     AD & .846 & .680 & .813 (6.1\% ${\color{red}\downarrow}$) & .570 (16.\% ${\color{red}\downarrow}$) \\
     RML+FC & .777 & .682 & .560 (28.\% ${\color{red}\downarrow}$)& .645 (5.4\% ${\color{red}\downarrow}$) \\
     RML+LBC & .866 & .671 & .869 (0.4\% ${\color{blue}\uparrow}$)& .646 (3.7\% ${\color{red}\downarrow}$)  \\
     RML+AD & .846 & .680 & .820 (3.1\% ${\color{red}\downarrow}$) & .573 (16.\% ${\color{red}\downarrow}$) \\
    \cmidrule(l){1-5}
     LR  & .465 & .674 & .454 (5.0\% ${\color{red}\downarrow}$) & .631 (6.4\% ${\color{red}\downarrow}$) \\
     RML & .493 & .680 & .575 (17.\% ${\color{blue}\uparrow}$) & .646 (5.0\% ${\color{red}\downarrow}$)   \\
     \cmidrule(l){1-5} 
     \textbf{\frtrain{}} & \textbf{.838} & \textbf{.676} & \textbf{.846 (1.0\% ${\color{blue}\uparrow}$)} & \textbf{.670 (0.9\% ${\color{red}\downarrow}$)}  \\
    
    \bottomrule
  \end{tabular}
  \vspace{-0.5cm}
\end{table}

\begin{table}[h]
  \vspace{-0.3cm}
  \caption{Accuracy and fairness results on AdultCensus test data w.r.t.\@ disparate impact (DI). Other settings are identical to Table~\ref{tbl:realCompas}.}
  \label{tbl:realAdult}
  \centering
  \begin{tabular}{l@{\hspace{7pt}}c@{\hspace{9pt}}c@{\hspace{12pt}}c@{\hspace{7pt}}c}
    \toprule
     Method & \multicolumn{2}{c}{Clean data} & \multicolumn{2}{c}{Poisoned data}                    \\
    \cmidrule(r){1-5}
      & DI & Acc. & DI & Acc. \\
    \cmidrule(r){1-5}   
    FC & .825 & .826 & .741 (10.\% ${\color{red}\downarrow}$)& .801 (3.0\% ${\color{red}\downarrow}$)  \\
    LBC & .825 & .825 & .760 (7.9\% ${\color{red}\downarrow}$)& .792 (4.0\% ${\color{red}\downarrow}$) \\
    AD & .850 & .767 & .755 (11.\% ${\color{red}\downarrow}$) & .563 (27.\% ${\color{red}\downarrow}$) \\
    RML+FC & .825 & .826 & .821 (0.5\% ${\color{red}\downarrow}$) & .780 (5.6\% ${\color{red}\downarrow}$)  \\
    RML+LBC & .825 & .825 & .762 (7.6\% ${\color{red}\downarrow}$) & .788 (4.5\% ${\color{red}\downarrow}$) \\
    RML+AD & .850 & .767 & .834 (1.9\% ${\color{red}\downarrow}$) & .647 (16.\% ${\color{red}\downarrow}$) \\
    \cmidrule(l){1-5}
    LR  & .328 & .847 & .189 (42.\% ${\color{red}\downarrow}$) & .819 (3.3\% ${\color{red}\downarrow}$)   \\
    RML  & .327 & .846 & .268 (18.\% ${\color{red}\downarrow}$) & .840 (0.7\% ${\color{red}\downarrow}$)   \\
     \cmidrule(l){1-5} 
    \textbf{\frtrain{}} & \textbf{.828} & \textbf{.824} & \textbf{.847 (2.3\% ${\color{blue}\uparrow}$)}& \textbf{.809 (1.8\% ${\color{red}\downarrow}$)}  \\
    
    \bottomrule
  \end{tabular}
\end{table}

\begin{table}[ht]
  \caption{Confusion matrix on poisoned AdultCensus dataset w.r.t.\@ disparate impact. Other settings are identical to Table~\ref{tbl:realCompas}.} 
  \label{tbl:ad_frgan_di}
  \centering
  \begin{tabular}{ll|cc|cc}
    \toprule
      Method& & \multicolumn{2}{c|}{Female} & \multicolumn{2}{c}{Male}  \\
    \cmidrule(r){1-6}
       & & $\hat{y}$ = 0 & $\hat{y}$ = 1 & $\hat{y}$ = 0 & $\hat{y}$ = 1  \\
    \midrule
    \multirow{2}{*}{RML+FC} & $y$  = 0 & 2,990 & 78 & 4,842 & 71 \\
    & $y$ = 1 & 238 & 147 & 1,952 & 313\\
    \midrule
    \multirow{2}{*}{RML+AD} & $y$  = 0 & 2,345 & 723 & 3,966 & 947 \\
    & $y$ = 1 & 289 & 96 & 1,792 & 473\\
    \midrule
    \multirow{2}{*}{\frtrain{}} & $y$  = 0 & 2,761 & 307 & 4,730 & 183 \\
    & $y$ = 1 & 113 & 272 & 1,428 & 837\\
    \bottomrule
  \end{tabular}
   \vspace{-0.35cm}
\end{table}

Table~\ref{tbl:ad_frgan_di} shows the confusion matrix comparison for disparate impact between \frtrain{}, AD, and FC with sanitization using RML, using the poisoned AdultCensus dataset. The results are reported when \frtrain{}, AD, and FC achieve (Acc, DI) = (0.809, 0.847), (0.647, 0.834), and (0.780, 0.821), respectively. FR-Train outperforms AD and FC in all aspects because its robustness discriminator is more effective in sanitizing poisoned data.

\subsection{Ablation Study}\label{ablation}
In Table~\ref{tbl:ablationCompas}, we perform an ablation study to investigate the effect of each component of \frtrain{}.
\textbf{(Without `R')}
When $\lambda_2 = 0$ (i.e., no robust training), disparate impact is high, but accuracy is low on the poisoned data, just like the other fairness-only methods (Table~\ref{tbl:realCompas}, rows 1--3). 
\textbf{(Without `F')} 
On the other hand, when $\lambda_1 = 0$ (i.e., no fair training), the accuracy is high, but the disparate impact is low, just like the other non-fairness methods (Table~\ref{tbl:realCompas}, rows 7--8).
\textbf{(Without Re-weighting)}
Finally, when not using example re-weighting, both accuracy and disparate impact are similar to or worse than \frtrain{}.

In summary, {\em only a holistic} framework like \frtrain{} {\em can achieve both} excellent model fairness and training robustness. In comparison, other methods tailored for only one of these objectives lose either accuracy, fairness or both.

\begin{table}[t]
  \caption{Ablation study for \frtrain{} on COMPAS test data w.r.t.\@ disparate impact (DI) where the training data is poisoned using the label flipping attack. Four methods are compared: (1) \frtrain{} without R ($\lambda_2 = 0$), (2) \frtrain{} without F ($\lambda_1 = 0$), (3) \frtrain{} without example re-weighting (Without RW), and (4) \frtrain{}. For rows 2--4, the validation set is 5\% of $\mathcal{D}_{tr}$.}
  \label{tbl:ablationCompas}
  \centering
  \begin{tabular}{l@{\hspace{3pt}}c@{\hspace{7pt}}c@{\hspace{10pt}}c@{\hspace{5pt}}c}
    \toprule
     Method & \multicolumn{2}{c}{Clean data} & \multicolumn{2}{c}{Poisoned data}                    \\
    \cmidrule(r){1-5}
      & DI & Acc. & DI & Acc. \\
    \midrule
     Without R & .846 & .678 & .802 (5.2\% ${\color{red}\downarrow}$) & .580 (14.\% ${\color{red}\downarrow}$) \\
     Without F & .482 & .681 & .420 (13.\% ${\color{red}\downarrow}$) & .632 (7.2\% ${\color{red}\downarrow}$) \\
     Without RW & .832 & .677 & .840 (1.0\% ${\color{blue}\uparrow}$) & .624 (7.8\% ${\color{red}\downarrow}$) \\
     \textbf{\frtrain{}} & \textbf{.838} & \textbf{.676} & \textbf{.846 (1.0\% ${\color{blue}\uparrow}$)} & \textbf{.670 (0.9\% ${\color{red}\downarrow}$)} \\
    \bottomrule
  \end{tabular}
  \vspace{-0.1cm}
\end{table}

\subsection{Error range of \frtrain{}}\label{sec:errorrange}
We investigate the error range of \frtrain{}. All the FR-Train experiments on the poisoned data are re-conducted with ten different random seeds to generate error ranges with mean ($m$) and standard deviation ($s$) values. The performances are reported in the form of $m\pm s/2$ in Table~\ref{tbl:errorrange}. On the synthetic and AdultCensus datasets, the lowest performances (i.e., $m - s/2$) of FR-Train are still better than the second-best performances in Tables~\ref{tbl:fairnesstradeoff} and~\ref{tbl:realAdult}, respectively. For the COMPAS dataset, the lowest performance of \frtrain{} is slightly worse than those of the LBC-related algorithms, which can be explained by the fact that the LBC algorithms were not affected much by the poisoning in the first place.

\begin{table}[htpb]
  \caption{Error range of \frtrain{} on the poisoned datasets w.r.t.\@ disparate impact (DI). The poisoned settings are identical to the previous experiments.}
  \label{tbl:errorrange}
  \centering
  \begin{tabular}{lcc}
    \toprule
     Dataset & \multicolumn{2}{c}{Poisoned data}\\
    \cmidrule(r){1-3}
      & DI & Acc.\\
    \cmidrule(r){1-3}   
    Synthetic & $0.795 \pm 0.019$ & $ 0.805 \pm 0.008$ \\
    COMPAS & $0.827 \pm 0.027$ & $0.653 \pm 0.005$ \\
    AdultCensus & $0.871 \pm 0.034$ & $0.796 \pm 0.006$ \\
    \bottomrule
  \end{tabular}
  \vspace{-0.1cm}
\end{table}

\subsection{Constructing a Clean Validation Set}\label{sec:crowd}
We now demonstrate how to construct a clean validation set using crowdsourcing. We construct validation sets for the COMPAS and AdultCensus datasets using Amazon Mechanical Turk (AMT). 
Although these datasets have labels, we assume that they are not available to use as clean data. 
We also release the datasets as a community resource (see the supplementary for the description and data) and believe our construction can be generalized to other datasets. 
While crowdsourcing is not the only way to construct a clean validation set, it is sufficient for our purposes.

We design the AMT task for each dataset by asking a worker to classify each example. For the AdultCensus dataset, a worker looks at various attributes of a person and predicts if a person has an income of at least \$50K. Instead of a yes/no answer, the answer must be on a scale of 1 to 4, which reflects the worker's opinion more accurately. The COMPAS dataset has a similar setting where the only difference is that the workers need to predict if a criminal will reoffend in two years. Each task displays about 30 questions where we pay 3 cents per answer. For quality control, each task also contains quizzes to educate the workers, and some questions are used to evaluate the performance of the workers. After collecting answers, we filter out poor performers, take the average of at most a fixed number of $N$ responses per question, and compare with the threshold 2.5 to produce the final labels. The number of answers per question can be fewer than $N$ if inaccurate workers are filtered out.
We used workers of all demographics in the US, Canada, and UK.
While this majority voting approach already works well in our experiments, one could additionally apply various quality control techniques like peer-reviewing that are known to further reduce bias~\cite{NIPS2011_4396}.

The important questions are how accurate the crowdsourced labels are and whether the constructed validation set results in high accuracy and fairness for \frtrain{}. Table~\ref{tbl:amt} shows the crowdsourced labels accuracies when $N$ increases from 1 to 11. Even for the highest accuracies, the predictions are not perfect because the workers are looking at limited information (i.e., only the features) without any other context. To see if the workers can do better, we also train logistic regression models on ground truth labels and show their accuracies on test data as upperbounds. As a result, the accuracies are comparable when $N$ = 11 for both the COMPAS and AdultCensus datasets. We thus use this setting for all experiments. Table~\ref{tbl:source of val set} shows how useful our constructed validation set is compared to using a ``perfect'' validation set of the same size made of ground truth labels. For both datasets, using a ground truth validation set results in slightly higher, but comparable disparate impacts while obtaining near-identical accuracies, justifying the use of crowdsourced validation sets for FR-Train.

\begin{table}[t]
  \caption{Accuracy comparison of the crowdsourced labels ($N$: number of answers averaged per example) and predictions of a logistic regression model trained on ground truth labels.}
  \label{tbl:amt}
  \centering
  \begin{tabular}{l@{\hspace{6pt}}c@{\hspace{6pt}}c@{\hspace{6pt}}c@{\hspace{6pt}}c}
    \toprule
    Dataset & \multicolumn{3}{c}{Crowdsourcing} & Trained Model \\
    \cmidrule(r){1-5}
      & $N$ = 1 & $N$ = 5 & $N$ = 11 & \\
    \midrule
    COMPAS & 0.609 & 0.656 & 0.667 & 0.659 \\
    AdultCensus & 0.645 & 0.721 & 0.743 & 0.804 \\
    \bottomrule
  \end{tabular}
  \vspace{-0.4cm}
\end{table}

\begin{table}[t]
  \caption{Accuracy and fairness of \frtrain{} when using crowdsourced labels versus ground truth labels for the validation set. The training data is poisoned as in Tables~\ref{tbl:realCompas} and \ref{tbl:realAdult}.}
  \label{tbl:source of val set}
  \centering
  \begin{tabular}{llcc}
    \toprule
    Dataset & Validation set & DI & Acc.  \\
    \midrule
     \multirow{2}{*}{COMPAS} &
     {Crowdsourcing}  & 0.846 & 0.670 \\
     &{Ground truth}  & 0.899 & 0.674 \\
     \midrule
     \multirow{2}{*}{AdultCensus} &
     {Crowdsourcing}  & 0.847 & 0.809 \\
     &{Ground truth}  & 0.864 & 0.809 \\
    \bottomrule
  \end{tabular}
  \vspace{-0.4cm}
\end{table}

\section{Related Work}
\label{sec:relatedwork}

\paragraph{Model Fairness}

The notion of discrimination has many definitions and usually comes from certain social goals that one wants to guarantee. As a result, many fairness measures have been proposed~\cite{DBLP:conf/icse/VermaR18}. 
While we focus on group fairness, which ensures similar statistics between two sensitive groups, an interesting future work is to consider individual fairness~\cite{Dwork:2012:FTA:2090236.2090255}, which guarantees similar prediction results across nearby examples.
Recently, there has also been a surge of research on unfairness mitigation techniques~\cite{DBLP:journals/corr/abs-1810-01943}. Depending on where a fix occurs, there are mainly three approaches: (1) {\em pre}-processing techniques~\cite{DBLP:journals/kais/KamiranC11,DBLP:conf/nips/CalmonWVRV17,DBLP:conf/icml/ZemelWSPD13,DBLP:conf/kdd/FeldmanFMSV15} that fix the training data; (2) {\em in}-processing techniques~\cite{DBLP:conf/aistats/ZafarVGG17,DBLP:journals/corr/abs-1901-04966,DBLP:conf/aies/ZhangLM18,DBLP:conf/pkdd/KamishimaAAS12,DBLP:conf/alt/CotterJS19,DBLP:journals/corr/abs-1807-00028,DBLP:conf/icml/AgarwalBD0W18} that address the issue during model training; and (3) {\em post}-processing techniques~\cite{DBLP:conf/nips/HardtPNS16,DBLP:conf/nips/PleissRWKW17,DBLP:conf/icdm/KamiranKZ12, NIPS2019_9437} that manipulate predictions while maintaining the model. Among the three, the in-processing techniques have the advantages that one can work with any data and that there is more control on model training~\cite{DBLP:conf/pods/Venkatasubramanian19}. 

Although not our immediate focus, there are other noteworthy directions in fairness research. Causality-based fairness~\cite{10.5555/3294771.3294834, NIPS2017_6995, Zhang2018FairnessID, Nabi2018FairIO, 10.1145/3308558.3313559, DBLP:conf/aaai/KhademiH20} suggests how to understand the causal relationship between attributes to overcome the limitations of non-causal approaches. Just as non-causal fairness can be captured by mutual information, we suspect there may be a connection between causal fairness and directed information.
Another important approach~\cite{pmlr-v80-hashimoto18a} is based on distributionally robust optimization (DRO)~\cite{Sinha2017CertifyingSD}, which focuses on when the sensitive attribute $z$ is unknown. The DRO-based fairness approach ensures fair results by equalizing risks over all distributions without the knowledge of $z$, but it does not directly minimize the fairness metrics such as disparate impact and equalized odds. In comparison, \frtrain{} assumes full knowledge of $z$ and utilizes it to directly minimize the fairness metrics.

As we demonstrate in Section~\ref{sec:vulnerability}, the existing fairness techniques are not tailored for robust training, so they are vulnerable to data poisoning attacks. In comparison, \frtrain{} addresses both model fairness and robust training within the same model training process because they are closely related and affected by the same training data. 
\vspace{-0.4cm}

\paragraph{Robust Training}

There is a heavy literature on how to make the model training robust against noisy or even adversarial data~\cite{DBLP:conf/nips/NatarajanDRT13,DBLP:journals/jmlr/BiggioNL11,DBLP:journals/tnn/FrenayV14,DBLP:conf/iclr/KurakinGB17}. A major challenge is that there can be a wide range of data poisoning attacks that keep on evolving. While sanitizing the training data before model training is an option, defending against all possible attacks seems fundamentally infeasible as demonstrated by~\cite{DBLP:journals/corr/abs-1811-00741}. A more recent trend is to develop general defense algorithms for any attack {\em during model training} using meta learning~\cite{DBLP:conf/cvpr/VeitACKGB17,DBLP:conf/iccv/LiYSCLL17,DBLP:conf/cvpr/XiaoXYHW15,DBLP:conf/nips/HendrycksMWG18}. Our \frtrain{} framework is inspired by robustness training with meta learning~\cite{DBLP:conf/icml/RenZYU18}, but employs a GAN-based model to support fair and robust training without using meta learning. In particular, the design of \frtrain{}'s robustness discriminator is based on mutual-information-based theoretical insights (Section~\ref{sec:robustness}). 
Another line of research is defending against adversarial attacks during {\em test} time~\cite{DBLP:conf/pkdd/BiggioCMNSLGR13,DBLP:journals/corr/GoodfellowSS14,DBLP:conf/icml/WongK18}. In comparison, our focus is on defending against data poisoning on the {\em training} data.

\section{Conclusion}

We proposed \frtrain{}, which is a holistic framework for trustworthy AI by performing both unfairness mitigation and robust training. Our key contribution is providing interpretation of an adversarial learning approach using mutual information and proposing a novel GAN architecture that enjoys the {\em synergistic effect} of combining two approaches: (1) employing a fairness discriminator that distinguishes predictions  w.r.t.\@ one sensitive group from others and (2) employing a robustness discriminator that distinguishes training data with predictions from a clean validation set and is also used to further improve the fairness training through example re-weighting. In addition, we demonstrated how a clean validation set can be constructed using crowdsourcing and released two new datasets built from Amazon Mechanical Turk as a community resource. In our experiments, we showed that existing fairness methods are vulnerable to data poisoning, even when combined with data sanitization. In comparison, \frtrain{} is robust to the poisoning and can be adjusted to maintain reasonable accuracy and fairness even if the validation set is too small or unavailable. 

\section*{Acknowledgements}
Yuji Roh and Steven E. Whang were supported by a Google AI Focused Research Award and by the Engineering Research Center Program through the National Research Foundation of Korea (NRF) funded by the Korean Government MSIT (NRF-2018R1A5A1059921).
This material is based upon work supported by the Air Force Office of Scientific Research under award number FA2386-19-1-4050.

\bibliography{main}

\begin{thebibliography}{54}
\providecommand{\natexlab}[1]{#1}
\providecommand{\url}[1]{\texttt{#1}}
\expandafter\ifx\csname urlstyle\endcsname\relax
  \providecommand{\doi}[1]{doi: #1}\else
  \providecommand{\doi}{doi: \begingroup \urlstyle{rm}\Url}\fi

\bibitem[Agarwal et~al.(2018)Agarwal, Beygelzimer, Dud{\'{\i}}k, Langford, and
  Wallach]{DBLP:conf/icml/AgarwalBD0W18}
Agarwal, A., Beygelzimer, A., Dud{\'{\i}}k, M., Langford, J., and Wallach,
  H.~M.
\newblock A reductions approach to fair classification.
\newblock In \emph{ICML}, pp.\  60--69, 2018.

\bibitem[Angwin et~al.(2016)Angwin, Larson, Mattu, and Kirchner]{Compas}
Angwin, J., Larson, J., Mattu, S., and Kirchner, L.
\newblock Machine bias: {T}here's software used across the country to predict
  future criminals. {A}nd its biased against blacks., 2016.

\bibitem[Bellamy et~al.(2018{\natexlab{a}})Bellamy, Dey, Hind, Hoffman, Houde,
  Kannan, Lohia, Martino, Mehta, Mojsilovic, Nagar, Ramamurthy, Richards, Saha,
  Sattigeri, Singh, Varshney, and Zhang]{aif360-oct-2018}
Bellamy, R. K.~E., Dey, K., Hind, M., Hoffman, S.~C., Houde, S., Kannan, K.,
  Lohia, P., Martino, J., Mehta, S., Mojsilovic, A., Nagar, S., Ramamurthy,
  K.~N., Richards, J., Saha, D., Sattigeri, P., Singh, M., Varshney, K.~R., and
  Zhang, Y.
\newblock {AI Fairness} 360: An extensible toolkit for detecting,
  understanding, and mitigating unwanted algorithmic bias, October
  2018{\natexlab{a}}.
\newblock URL \url{https://arxiv.org/abs/1810.01943}.

\bibitem[Bellamy et~al.(2018{\natexlab{b}})Bellamy, Dey, Hind, Hoffman, Houde,
  Kannan, Lohia, Martino, Mehta, Mojsilovic, Nagar, Ramamurthy, Richards, Saha,
  Sattigeri, Singh, Varshney, and Zhang]{DBLP:journals/corr/abs-1810-01943}
Bellamy, R. K.~E., Dey, K., Hind, M., Hoffman, S.~C., Houde, S., Kannan, K.,
  Lohia, P., Martino, J., Mehta, S., Mojsilovic, A., Nagar, S., Ramamurthy,
  K.~N., Richards, J.~T., Saha, D., Sattigeri, P., Singh, M., Varshney, K.~R.,
  and Zhang, Y.
\newblock {AI} fairness 360: {A}n extensible toolkit for detecting,
  understanding, and mitigating unwanted algorithmic bias.
\newblock \emph{CoRR}, abs/1810.01943, 2018{\natexlab{b}}.

\bibitem[Biggio et~al.(2011)Biggio, Nelson, and
  Laskov]{DBLP:journals/jmlr/BiggioNL11}
Biggio, B., Nelson, B., and Laskov, P.
\newblock Support vector machines under adversarial label noise.
\newblock In \emph{ACML}, pp.\  97--112, 2011.

\bibitem[Biggio et~al.(2013)Biggio, Corona, Maiorca, Nelson, Srndic, Laskov,
  Giacinto, and Roli]{DBLP:conf/pkdd/BiggioCMNSLGR13}
Biggio, B., Corona, I., Maiorca, D., Nelson, B., Srndic, N., Laskov, P.,
  Giacinto, G., and Roli, F.
\newblock Evasion attacks against machine learning at test time.
\newblock In \emph{ECML PKDD}, pp.\  387--402, 2013.
\newblock \doi{10.1007/978-3-642-40994-3\_25}.

\bibitem[Chouldechova \& Roth(2018)Chouldechova and
  Roth]{DBLP:journals/corr/abs-1810-08810}
Chouldechova, A. and Roth, A.
\newblock The frontiers of fairness in machine learning.
\newblock \emph{CoRR}, abs/1810.08810, 2018.

\bibitem[Chzhen et~al.(2019)Chzhen, Denis, Hebiri, Oneto, and
  Pontil]{NIPS2019_9437}
Chzhen, E., Denis, C., Hebiri, M., Oneto, L., and Pontil, M.
\newblock Leveraging labeled and unlabeled data for consistent fair binary
  classification.
\newblock In \emph{NeurIPS}, pp.\  12760--12770. 2019.

\bibitem[Cotter et~al.(2018)Cotter, Gupta, Jiang, Srebro, Sridharan, Wang,
  Woodworth, and You]{DBLP:journals/corr/abs-1807-00028}
Cotter, A., Gupta, M.~R., Jiang, H., Srebro, N., Sridharan, K., Wang, S.,
  Woodworth, B.~E., and You, S.
\newblock Training well-generalizing classifiers for fairness metrics and other
  data-dependent constraints.
\newblock \emph{CoRR}, abs/1807.00028, 2018.

\bibitem[Cotter et~al.(2019)Cotter, Jiang, and
  Sridharan]{DBLP:conf/alt/CotterJS19}
Cotter, A., Jiang, H., and Sridharan, K.
\newblock Two-player games for efficient non-convex constrained optimization.
\newblock In \emph{ALT}, pp.\  300--332, 2019.

\bibitem[du~Pin~Calmon et~al.(2017)du~Pin~Calmon, Wei, Vinzamuri, Ramamurthy,
  and Varshney]{DBLP:conf/nips/CalmonWVRV17}
du~Pin~Calmon, F., Wei, D., Vinzamuri, B., Ramamurthy, K.~N., and Varshney,
  K.~R.
\newblock Optimized pre-processing for discrimination prevention.
\newblock In \emph{NeurIPS}, pp.\  3995--4004, 2017.

\bibitem[Dwork et~al.(2012)Dwork, Hardt, Pitassi, Reingold, and
  Zemel]{Dwork:2012:FTA:2090236.2090255}
Dwork, C., Hardt, M., Pitassi, T., Reingold, O., and Zemel, R.
\newblock Fairness through awareness.
\newblock In \emph{ITCS}, pp.\  214--226, 2012.
\newblock ISBN 978-1-4503-1115-1.

\bibitem[Feldman et~al.(2015)Feldman, Friedler, Moeller, Scheidegger, and
  Venkatasubramanian]{DBLP:conf/kdd/FeldmanFMSV15}
Feldman, M., Friedler, S.~A., Moeller, J., Scheidegger, C., and
  Venkatasubramanian, S.
\newblock Certifying and removing disparate impact.
\newblock In \emph{KDD}, pp.\  259--268, 2015.
\newblock \doi{10.1145/2783258.2783311}.

\bibitem[Fr{\'{e}}nay \& Verleysen(2014)Fr{\'{e}}nay and
  Verleysen]{DBLP:journals/tnn/FrenayV14}
Fr{\'{e}}nay, B. and Verleysen, M.
\newblock Classification in the presence of label noise: {A} survey.
\newblock \emph{{IEEE} Trans. Neural Netw. Learning Syst.}, 25\penalty0
  (5):\penalty0 845--869, 2014.
\newblock \doi{10.1109/TNNLS.2013.2292894}.

\bibitem[Goodfellow et~al.(2014)Goodfellow, Pouget{-}Abadie, Mirza, Xu,
  Warde{-}Farley, Ozair, Courville, and
  Bengio]{DBLP:conf/nips/GoodfellowPMXWOCB14}
Goodfellow, I.~J., Pouget{-}Abadie, J., Mirza, M., Xu, B., Warde{-}Farley, D.,
  Ozair, S., Courville, A.~C., and Bengio, Y.
\newblock Generative adversarial nets.
\newblock In \emph{NeurIPS}, pp.\  2672--2680, 2014.

\bibitem[Goodfellow et~al.(2015)Goodfellow, Shlens, and
  Szegedy]{DBLP:journals/corr/GoodfellowSS14}
Goodfellow, I.~J., Shlens, J., and Szegedy, C.
\newblock Explaining and harnessing adversarial examples.
\newblock In \emph{ICLR}, 2015.

\bibitem[Hardt et~al.(2016)Hardt, Price, and Srebro]{DBLP:conf/nips/HardtPNS16}
Hardt, M., Price, E., and Srebro, N.
\newblock Equality of opportunity in supervised learning.
\newblock In \emph{NeurIPS}, pp.\  3315--3323, 2016.

\bibitem[Hashimoto et~al.(2018)Hashimoto, Srivastava, Namkoong, and
  Liang]{pmlr-v80-hashimoto18a}
Hashimoto, T., Srivastava, M., Namkoong, H., and Liang, P.
\newblock Fairness without demographics in repeated loss minimization.
\newblock In \emph{ICML}, pp.\  1929--1938, 2018.

\bibitem[Hendrycks et~al.(2018)Hendrycks, Mazeika, Wilson, and
  Gimpel]{DBLP:conf/nips/HendrycksMWG18}
Hendrycks, D., Mazeika, M., Wilson, D., and Gimpel, K.
\newblock Using trusted data to train deep networks on labels corrupted by
  severe noise.
\newblock In \emph{NeurIPS}, pp.\  10477--10486, 2018.

\bibitem[IBM(2020)]{trustedai}
IBM.
\newblock Trusting ai.
\newblock
  \url{https://www.research.ibm.com/artificial-intelligence/trusted-ai/}, 2020.

\bibitem[Jiang \& Nachum(2019)Jiang and
  Nachum]{DBLP:journals/corr/abs-1901-04966}
Jiang, H. and Nachum, O.
\newblock Identifying and correcting label bias in machine learning.
\newblock \emph{CoRR}, abs/1901.04966, 2019.

\bibitem[Kamiran \& Calders(2011)Kamiran and
  Calders]{DBLP:journals/kais/KamiranC11}
Kamiran, F. and Calders, T.
\newblock Data preprocessing techniques for classification without
  discrimination.
\newblock \emph{Knowl. Inf. Syst.}, 33\penalty0 (1):\penalty0 1--33, 2011.
\newblock \doi{10.1007/s10115-011-0463-8}.

\bibitem[Kamiran et~al.(2012)Kamiran, Karim, and
  Zhang]{DBLP:conf/icdm/KamiranKZ12}
Kamiran, F., Karim, A., and Zhang, X.
\newblock Decision theory for discrimination-aware classification.
\newblock In \emph{ICDM}, pp.\  924--929, 2012.
\newblock \doi{10.1109/ICDM.2012.45}.

\bibitem[Kamishima et~al.(2012)Kamishima, Akaho, Asoh, and
  Sakuma]{DBLP:conf/pkdd/KamishimaAAS12}
Kamishima, T., Akaho, S., Asoh, H., and Sakuma, J.
\newblock Fairness-aware classifier with prejudice remover regularizer.
\newblock In \emph{ECML PKDD}, pp.\  35--50, 2012.
\newblock \doi{10.1007/978-3-642-33486-3\_3}.

\bibitem[Karger et~al.(2011)Karger, Oh, and Shah]{NIPS2011_4396}
Karger, D.~R., Oh, S., and Shah, D.
\newblock Iterative learning for reliable crowdsourcing systems.
\newblock In \emph{NIPS}, pp.\  1953--1961. 2011.

\bibitem[Karpathy(2017)]{karpathi}
Karpathy, A.
\newblock Software 2.0.
\newblock \url{https://medium.com/@karpathy/software-2-0-a64152b37c35}, 2017.

\bibitem[Khademi \& Honavar(2020)Khademi and
  Honavar]{DBLP:conf/aaai/KhademiH20}
Khademi, A. and Honavar, V.~G.
\newblock Algorithmic bias in recidivism prediction: {A} causal perspective
  (student abstract).
\newblock In \emph{AAAI}, pp.\  13839--13840, 2020.

\bibitem[Khademi et~al.(2019)Khademi, Lee, Foley, and
  Honavar]{10.1145/3308558.3313559}
Khademi, A., Lee, S., Foley, D., and Honavar, V.
\newblock Fairness in algorithmic decision making: An excursion through the
  lens of causality.
\newblock In \emph{WWW}, pp.\  2907–--2914, 2019.

\bibitem[Kilbertus et~al.(2017)Kilbertus, Rojas-Carulla, Parascandolo, Hardt,
  Janzing, and Sch\"{o}lkopf]{10.5555/3294771.3294834}
Kilbertus, N., Rojas-Carulla, M., Parascandolo, G., Hardt, M., Janzing, D., and
  Sch\"{o}lkopf, B.
\newblock Avoiding discrimination through causal reasoning.
\newblock In \emph{NeurIPS}, pp.\  656–--666, 2017.

\bibitem[Kingma \& Ba(2014)Kingma and Ba]{kingma2014adam}
Kingma, D.~P. and Ba, J.
\newblock Adam: {A} method for stochastic optimization.
\newblock \emph{arXiv preprint arXiv:1412.6980}, 2014.

\bibitem[Koh et~al.(2018)Koh, Steinhardt, and
  Liang]{DBLP:journals/corr/abs-1811-00741}
Koh, P.~W., Steinhardt, J., and Liang, P.
\newblock Stronger data poisoning attacks break data sanitization defenses.
\newblock \emph{CoRR}, abs/1811.00741, 2018.

\bibitem[Kohavi(1996)]{DBLP:conf/kdd/Kohavi96}
Kohavi, R.
\newblock Scaling up the accuracy of naive-bayes classifiers: {A} decision-tree
  hybrid.
\newblock In \emph{KDD}, pp.\  202--207, 1996.

\bibitem[Kurakin et~al.(2017)Kurakin, Goodfellow, and
  Bengio]{DBLP:conf/iclr/KurakinGB17}
Kurakin, A., Goodfellow, I.~J., and Bengio, S.
\newblock Adversarial machine learning at scale.
\newblock In \emph{ICLR}, 2017.

\bibitem[Kusner et~al.(2017)Kusner, Loftus, Russell, and Silva]{NIPS2017_6995}
Kusner, M.~J., Loftus, J., Russell, C., and Silva, R.
\newblock Counterfactual fairness.
\newblock In \emph{NeurIPS}, pp.\  4066--4076. 2017.

\bibitem[Li et~al.(2017)Li, Yang, Song, Cao, Luo, and
  Li]{DBLP:conf/iccv/LiYSCLL17}
Li, Y., Yang, J., Song, Y., Cao, L., Luo, J., and Li, L.
\newblock Learning from noisy labels with distillation.
\newblock In \emph{ICCV}, pp.\  1928--1936, 2017.
\newblock \doi{10.1109/ICCV.2017.211}.

\bibitem[Lin(1991)]{lin1991divergence}
Lin, J.
\newblock Divergence measures based on the {S}hannon entropy.
\newblock \emph{IEEE Transactions on Information theory}, 37\penalty0
  (1):\penalty0 145--151, 1991.

\bibitem[Nabi \& Shpitser(2018)Nabi and Shpitser]{Nabi2018FairIO}
Nabi, R. and Shpitser, I.
\newblock Fair inference on outcomes.
\newblock \emph{AAAI}, pp.\  1931--1940, 2018.

\bibitem[Natarajan et~al.(2013)Natarajan, Dhillon, Ravikumar, and
  Tewari]{DBLP:conf/nips/NatarajanDRT13}
Natarajan, N., Dhillon, I.~S., Ravikumar, P., and Tewari, A.
\newblock Learning with noisy labels.
\newblock In \emph{NeurIPS}, pp.\  1196--1204, 2013.

\bibitem[Noy et~al.(2019)Noy, Burgess, and Brickley]{47845}
Noy, N., Burgess, M., and Brickley, D.
\newblock Google dataset search: Building a search engine for datasets in an
  open web ecosystem.
\newblock In \emph{28th Web Conference (WebConf 2019)}, 2019.

\bibitem[Paszke et~al.(2017)Paszke, Gross, Chintala, Chanan, Yang, DeVito, Lin,
  Desmaison, Antiga, and Lerer]{paszke2017automatic}
Paszke, A., Gross, S., Chintala, S., Chanan, G., Yang, E., DeVito, Z., Lin, Z.,
  Desmaison, A., Antiga, L., and Lerer, A.
\newblock Automatic differentiation in {PyTorch}.
\newblock In \emph{NIPS Autodiff Workshop}, 2017.

\bibitem[Paudice et~al.(2018)Paudice, Mu{\~{n}}oz{-}Gonz{\'{a}}lez, and
  Lupu]{DBLP:conf/pkdd/PaudiceML18}
Paudice, A., Mu{\~{n}}oz{-}Gonz{\'{a}}lez, L., and Lupu, E.~C.
\newblock Label sanitization against label flipping poisoning attacks.
\newblock In \emph{ECML PKDD}, pp.\  5--15, 2018.
\newblock \doi{10.1007/978-3-030-13453-2\_1}.

\bibitem[Pleiss et~al.(2017)Pleiss, Raghavan, Wu, Kleinberg, and
  Weinberger]{DBLP:conf/nips/PleissRWKW17}
Pleiss, G., Raghavan, M., Wu, F., Kleinberg, J.~M., and Weinberger, K.~Q.
\newblock On fairness and calibration.
\newblock In \emph{NeurIPS}, pp.\  5684--5693, 2017.

\bibitem[Ren et~al.(2018)Ren, Zeng, Yang, and Urtasun]{DBLP:conf/icml/RenZYU18}
Ren, M., Zeng, W., Yang, B., and Urtasun, R.
\newblock Learning to reweight examples for robust deep learning.
\newblock In \emph{ICML}, pp.\  4331--4340, 2018.

\bibitem[Sinha et~al.(2017)Sinha, Namkoong, and Duchi]{Sinha2017CertifyingSD}
Sinha, A., Namkoong, H., and Duchi, J.~C.
\newblock Certifying some distributional robustness with principled adversarial
  training.
\newblock In \emph{ICLR}, 2017.

\bibitem[Veit et~al.(2017)Veit, Alldrin, Chechik, Krasin, Gupta, and
  Belongie]{DBLP:conf/cvpr/VeitACKGB17}
Veit, A., Alldrin, N., Chechik, G., Krasin, I., Gupta, A., and Belongie, S.~J.
\newblock Learning from noisy large-scale datasets with minimal supervision.
\newblock In \emph{CVPR}, pp.\  6575--6583, 2017.
\newblock \doi{10.1109/CVPR.2017.696}.

\bibitem[Venkatasubramanian(2019)]{DBLP:conf/pods/Venkatasubramanian19}
Venkatasubramanian, S.
\newblock Algorithmic fairness: Measures, methods and representations.
\newblock In \emph{PODS}, pp.\  481, 2019.
\newblock \doi{10.1145/3294052.3322192}.

\bibitem[Verma \& Rubin(2018)Verma and Rubin]{DBLP:conf/icse/VermaR18}
Verma, S. and Rubin, J.
\newblock Fairness definitions explained.
\newblock In \emph{FairWare@ICSE}, pp.\  1--7, 2018.
\newblock \doi{10.1145/3194770.3194776}.

\bibitem[Wong \& Kolter(2018)Wong and Kolter]{DBLP:conf/icml/WongK18}
Wong, E. and Kolter, J.~Z.
\newblock Provable defenses against adversarial examples via the convex outer
  adversarial polytope.
\newblock In \emph{ICML}, pp.\  5283--5292, 2018.

\bibitem[Xiao et~al.(2015)Xiao, Xia, Yang, Huang, and
  Wang]{DBLP:conf/cvpr/XiaoXYHW15}
Xiao, T., Xia, T., Yang, Y., Huang, C., and Wang, X.
\newblock Learning from massive noisy labeled data for image classification.
\newblock In \emph{CVPR}, pp.\  2691--2699, 2015.
\newblock \doi{10.1109/CVPR.2015.7298885}.

\bibitem[Zafar et~al.(2017)Zafar, Valera, Gomez{-}Rodriguez, and
  Gummadi]{DBLP:conf/aistats/ZafarVGG17}
Zafar, M.~B., Valera, I., Gomez{-}Rodriguez, M., and Gummadi, K.~P.
\newblock Fairness constraints: {M}echanisms for fair classification.
\newblock In \emph{AISTATS}, pp.\  962--970, 2017.

\bibitem[Zemel et~al.(2013)Zemel, Wu, Swersky, Pitassi, and
  Dwork]{DBLP:conf/icml/ZemelWSPD13}
Zemel, R.~S., Wu, Y., Swersky, K., Pitassi, T., and Dwork, C.
\newblock Learning fair representations.
\newblock In \emph{ICML}, pp.\  325--333, 2013.

\bibitem[Zhang et~al.(2018{\natexlab{a}})Zhang, Lemoine, and
  Mitchell]{DBLP:conf/aies/ZhangLM18}
Zhang, B.~H., Lemoine, B., and Mitchell, M.
\newblock Mitigating unwanted biases with adversarial learning.
\newblock In \emph{AIES}, pp.\  335--340, 2018{\natexlab{a}}.
\newblock \doi{10.1145/3278721.3278779}.

\bibitem[Zhang \& Bareinboim(2018)Zhang and Bareinboim]{Zhang2018FairnessID}
Zhang, J. and Bareinboim, E.
\newblock Fairness in decision-making - the causal explanation formula.
\newblock In \emph{AAAI}, 2018.

\bibitem[Zhang et~al.(2018{\natexlab{b}})Zhang, Zhu, and
  Wright]{Zhang2018TrainingSD}
Zhang, X., Zhu, X., and Wright, S.~J.
\newblock Training set debugging using trusted items.
\newblock In \emph{AAAI}, 2018{\natexlab{b}}.

\end{thebibliography}
\bibliographystyle{icml2020}

\clearpage
\newpage
\appendix

\section{Appendix}

Appendix~\ref{sec:thm1proof} proves Theorem~\ref{thm:mi}. Appendix~\ref{sec:eo_extension} extends the theoretical results of the fairness discriminator to other measures. Appendix~\ref{sec:additionalexperiments} shows additional experiments. Appendix~\ref{sec:trainingmethodology} provides more details of the model training setup.

\subsection{Proof for Theorem~\ref{thm:mi}}
\label{sec:thm1proof}
Before we present the proof of the main theorem, we first recall our notation.
Let $P_Z(z)$ be the distribution of $Z$ where $z \in \mathcal{Z}$ and $\mathcal{Z}$ is the set of possible sensitive attribute values. Let $\hat{Y} | Z = z \sim P_{\hat{Y}|z}(\cdot)$ and $\hat{Y} \sim P_{\hat{Y}}(\cdot)$. Then $P_{\hat{Y}}(\cdot) = \sum_{z \in \mathcal{Z}} P_Z(z) P_{\hat{Y}|z}(\cdot)$. Also, let $Y \sim P_Y(\cdot).$

For convenience, let us repeat the statement of Theorem~\ref{thm:mi} here:
\begin{align*}
&I(Z;\hat{Y}) =\\
&\max_{D_z(\hat{y}):\sum_z D_z(\hat{y})=1,~\forall \hat{y}}~\sum_{z \in \mathcal{Z}}P_Z(z)\E_{P_{\hat{Y}|z}} \left[ \log D_z(\hat{Y}) \right] \\&+ H(Z).
\end{align*}
We now prove the theorem.
\begin{proof}

Denote by $\bm{D}$ the collection of $D_z(\hat{y})$ for all possible values of $z$ and $\hat{y}$, and by $\bm{\nu}$ the collection of $\nu_{\hat{y}}$ for all values of $\hat{y}$.
We can construct the Lagrangian function as follows:
\begin{align*}
    \mathcal{L}(\bm{D}, \bm{\nu}) =& \sum_{z \in \mathcal{Z}} P_Z(z)\E_{P_{\hat{Y}|z}} \left[ \log D_z(\hat{Y}) \right] + H(Z) \\
    &+ \sum_{\hat{y}\in\mathcal{Y}}\nu_{\hat{y}}  \left( 1 - \sum_{z \in \mathcal{Z}} D_z(\hat{y})\right).
\end{align*}

We use the following KKT conditions:
\begin{align*}
  &\frac{\partial \mathcal{L}(\bm{D}, \bm{\nu})}{\partial D_z(\hat{y})} = P_Z(z) \frac{P_{\hat{Y}|z} (\hat{y})}{D_z^{\star}(\hat{y})} - \nu^{\star}_{\hat{y}} = 0,&&\forall (\hat{y}, z) \in \mathcal{Y} \times \mathcal{Z}, \\
  &1 - \sum_{z \in \mathcal{Z}} D_z^{\star}(\hat{y}) = 0,&&\forall \hat{y} \in \mathcal{Y}.
\end{align*}

Solving the two equations, we obtain $\nu^{\star}_{\hat{y}} = P_{\hat{Y}} (\hat{y})$ for all $\hat{y}$. 
Thus,
\begin{align*}
    D_z^{\star}(\hat{y}) = \frac{P_Z(z)P_{\hat{Y}|z}(\hat{y})}{P_{\hat{Y}}(\hat{y})}.
\end{align*}
Putting this to the above optimization,
\begin{align*}
&\sum_{z \in \mathcal{Z}} P_Z(z) \E_{P_{\hat{Y}|z}} \left[ \log \frac{P_Z(z)P_{\hat{Y}|z}(\hat{Y})}{P_{\hat{Y}}(\hat{Y})}\right] + H(Z)\\
=&\sum_{z \in \mathcal{Z}} P_Z(z) \E_{P_{\hat{Y}|z}} \left[ \log \frac{P_Z(z)P_{\hat{Y}|z}(\hat{Y})}{P_{\hat{Y}}(\hat{Y})}\right] \\
&+ \sum_{z \in \mathcal{Z}}P_Z(z) \log \frac{1}{P_Z(z)}\\
=&\sum_{z \in \mathcal{Z}} P_Z(z) \E_{P_{\hat{Y}|z}} \left[ \log \frac{P_{\hat{Y}|z}(\hat{Y})}{P_{\hat{Y}}(\hat{Y})}\right]\\
=& \sum_{z \in \mathcal{Z}} P_Z(z) \KL(P_{\hat{Y}|z} \Vert P_{\hat{Y}})\nonumber\\
\triangleq& ~\mathrm{JS}_{P_Z}(P_{\hat{Y}|z_1}, \ldots, P_{\hat{Y}|z_{|\mathcal{Z}|}})\nonumber = I(Z;\hat{Y}).
\end{align*}
Here, the second last equality is due to the definition of the generalized Jensen-Shannon divergence, and the last equality is due to its equivalence to the mutual information~\cite{lin1991divergence}.\qedhere
\end{proof}

\subsection{Extensions to other fairness measures}
\label{sec:eo_extension}

We now extend \frtrain{} to the case of \emph{equalized odds}, which is another important fairness metric, defined as follows:

\begin{definition}{(Equalized Odds)}
\\$P(\hat{Y}=1|Y=y, Z=z_1) = P(\hat{Y}=1|Y=y, Z=z_2),$
\\$\forall y \in \mathcal{Y},~\forall z_1, z_2 \in \mathcal{Z}$. 
\end{definition}

The following theorem relates the conditional mutual information $I(Z;\hat{Y}|Y)$ to the solution of an optimization problem. 
\begin{thm}
\label{thm:cmi}
$I(Z;\hat{Y}|Y)$ =\\
$\max_{D_{z|y}(\hat{y}):\sum_{z \in \mathcal{Z}} D_{z|y}(\hat{y})=1,~\forall \hat{y}}$ \\
\resizebox{\hsize}{!}{$\sum_{y \in \mathcal{Y}}\sum_{z \in \mathcal{Z}}P_{Y,Z}(y,z)\E_{P_{\hat{Y}|y,z}} \left[ \log D_{z|y}(\hat{Y}) \right] + H(Z|Y).$}
\end{thm}

This conditional mutual information term can be used to capture \emph{equalized odds}.
We also note that the following theorem can be modified in a straightforward manner so that it can handle $I(Z;\hat{Y}|Y=1)$, which can be used to capture \emph{equal opportunity}.

We now prove the theorem.
\begin{proof}
Denote by $\bm{D}$ the collection of $D_{z|y}(\hat{y})$ for all possible values of ($z$, $\hat{y}$, and $y$) and by $\bm{\nu}$ the collection of $\nu_{y, \hat{y}}$ for all values of $y$ and $\hat{y}$.
We can construct the Lagrangian function as follows:
\begin{align*}
    \mathcal{L}(\bm{D}, \bm{\nu}) &= \sum_{y \in \mathcal{Y}}\sum_{z \in \mathcal{Z}} P_{Y,Z}(y,z)\E_{P_{\hat{Y}|y,z}} \left[ \log D_{z|y}(\hat{Y}) \right] \\
    &+ H(Z|Y) + \sum_{\hat{y} \in \mathcal{Y}} \sum_{y \in \mathcal{Y}} \nu_{y,\hat{y}} \left( 1 - \sum_{z \in \mathcal{Z}} D_{z|y}(\hat{y})\right).
\end{align*}

We use the following KKT conditions:
\begin{align*}
    &\frac{\partial \mathcal{L}(\bm{D}, \bm{\nu})}{\partial D_{z|y}(\hat{y})} = P_{Y,Z}(y,z) \frac{P_{\hat{Y}|y,z} (\hat{y})}{D_{z|y}^{\star}(\hat{y})} - \nu_{y,\hat{y}} = 0, \\
  &\hspace{5cm}\forall (\hat{y}, y, z) \in \mathcal{Y} \times \mathcal{Y} \times \mathcal{Z}\\
    &1 - \sum_{z \in \mathcal{Z}} D_{z|y}^{\star}(\hat{y}) = 0,~\forall (\hat{y}, y) \in \mathcal{Y} \times \mathcal{Y}.
\end{align*}
Solving the two equations, we obtain $\nu^{\star}_{y,\hat{y}} = P_{Y, \hat{Y}} (y, \hat{y})$ for all $(y, \hat{y}) \in \mathcal{Y} \times \mathcal{Y}$. 
Thus,
\begin{align*}
    D_{z|y}^{\star}(\hat{y}) = \frac{P_{Z|y}(z)P_{\hat{Y}|y,z}(\hat{y})}{P_{\hat{Y}|y}(\hat{y})},~\forall y,\hat{y} \in \mathcal{Y} \times \mathcal{Y}.
\end{align*}
Putting this to the above optimization,
\begin{align*}
&\sum_{y \in \mathcal{Y}} \sum_{z \in \mathcal{Z}}P_{Y,Z}(y,z)\E_{P_{\hat{Y}|y,z}} \left[ \log \frac{P_{Z|y}(z)P_{\hat{Y}|y,z}(\hat{Y})}{P_{\hat{Y}|y}(\hat{Y})}\right]\\
&+ H(Z|Y)\\
= &\sum_{y \in \mathcal{Y}} \sum_{z \in \mathcal{Z}}P_{Y,Z}(y,z)\E_{P_{\hat{Y}|y,z}} \left[ \log \frac{P_{Z|y}(z)P_{\hat{Y}|y,z}(\hat{Y})}{P_{\hat{Y}|y}(\hat{Y})}\right]\\
&+ \sum_{y \in \mathcal{Y}} \sum_{z \in \mathcal{Z}}P_{Y,Z}(y,z)\log \frac{1}{P_{Z|y}(z)} \\
= &\sum_{y \in \mathcal{Y}} \sum_{z \in \mathcal{Z}}P_{Y,Z}(y,z)\E_{P_{\hat{Y}|y,z}} \left[ \log \frac{P_{\hat{Y}|y,z}(\hat{Y})}{P_{\hat{Y}|y}(\hat{Y})}\right]\\
= & \sum_{y \in \mathcal{Y}} \sum_{z \in \mathcal{Z}}P_Y(y)P_{Z|y}(z)\E_{P_{\hat{Y}|y,z}} \left[ \log \frac{P_{\hat{Y}|y,z}(\hat{Y})}{P_{\hat{Y}|y}(\hat{Y})}\right]\\
= & \sum_{y \in \mathcal{Y}} P_Y(y) \sum_{z \in \mathcal{Z}}P_{Z|y}(z)\E_{P_{\hat{Y}|y,z}} \left[ \log \frac{P_{\hat{Y}|y,z}(\hat{Y})}{P_{\hat{Y}|y}(\hat{Y})}\right]\\
= & \sum_{y \in \mathcal{Y}} P_Y(y) \sum_{z \in \mathcal{Z}}P_{Z|y}(z) \KL(P_{\hat{Y}|y,z} \Vert P_{\hat{Y}|y})\\
\triangleq & \sum_{y \in \mathcal{Y}} P_Y(y) \cdot  \mathrm{JS}_{P_{Z|y}}(P_{\hat{Y}|z_1,y}, \ldots, P_{\hat{Y}|z_{|\mathcal{Z}|},y})\\
= & \sum_{y \in \mathcal{Y}}P_Y(y)I(Z;\hat{Y}|Y=y) = I(Z;\hat{Y}|Y).
\end{align*}
The third last equality is due to the definition of the generalized Jensen-Shannon divergence; the second last equality is due to its equivalence to the mutual information~\cite{lin1991divergence}; and the last equality is due to the definition of conditional mutual information. \qedhere
\end{proof}

We now discuss how to actually compute the mutual information. We compute the following empirical version using the examples \{($x^{(i)}$, $z^{(i)}$,  $y^{(i)}$)\}$_{i=1}^m$.
\begin{align*}
    &\max_{D_{z|y}(\hat{y}):\sum_{z \in \mathcal{Z}} D_{z|y}(\hat{y})=1; \forall \hat{y}} \sum_{y \in \mathcal{Y}} \sum_{z \in \mathcal{Z}} P_{Y,Z}(y,z)\\ &\hspace{1cm} \sum_{i:(y^{(i)}, z^{(i)})=(y,z)} \frac{1}{m_{y,z}} \log D_{z|y}(\hat{y}^{(i)}) + H(Z|Y).
\end{align*}
Now for a sufficiently large value of $m$, $m_{y,z} \approx P_{Y,Z}(y,z)m$. Therefore, the above expression is approximated as:
\begin{align*}
    &\max_{D_{z|y}(\hat{y}):\sum_{z \in \mathcal{Z}} D_{z|y}(\hat{y})=1; \forall \hat{y}}\sum_{y \in \mathcal{Y}}\sum_{z \in \mathcal{Z}}\\
    &\hspace{1cm} \sum_{i:(y^{(i)}, z^{(i)})=(y,z)} \frac{1}{m} \log D_{z|y}(\hat{y}^{(i)}) + H(Z|Y).
\end{align*}

\begin{figure*}[t]
  \centering
     \includegraphics[width=0.8\textwidth]{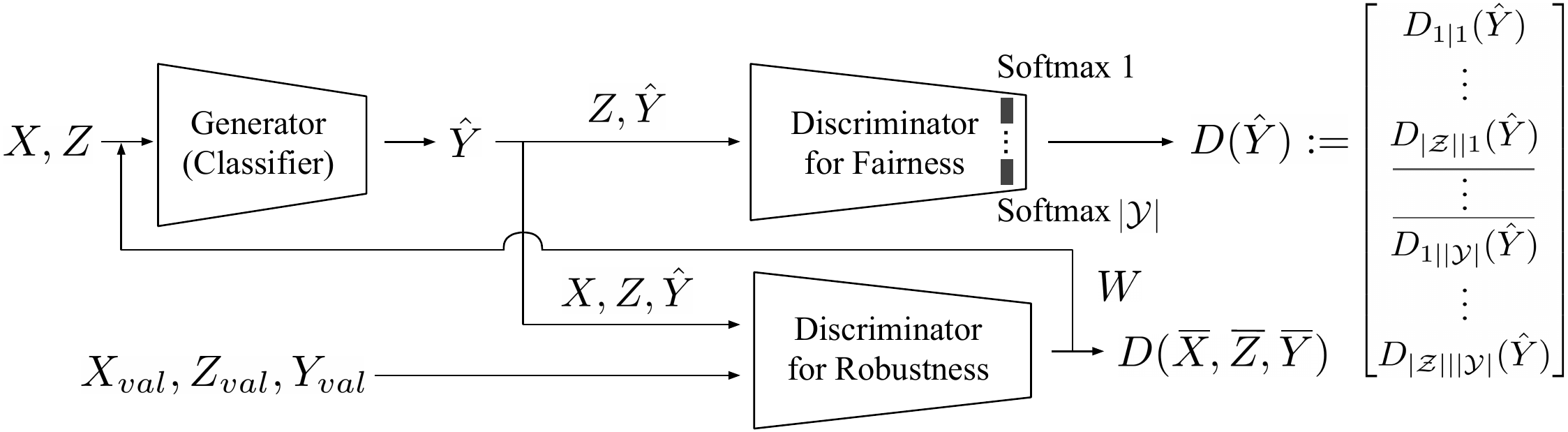}
     \caption{The architecture of \frtrain{} for equalized odds.}
 \label{fig:frgan_eo}
\end{figure*}

Hence, we can set $L_2$ (i.e., the loss w.r.t.\@ the fairness discriminator) to the above expression. The rest of the objective function is the same. Figure~\ref{fig:frgan_eo} shows the resulting \frtrain{} architecture.

\subsection{Additional experiments}
\label{sec:additionalexperiments}

\subsubsection{Synthetic data}
We continue our experiments from Section~\ref{sec:syntheticdataresults}. In particular, we perform \frtrain{} with different amounts of poisoning, and evaluate robust training with meta learning using smaller validation sets.

\paragraph*{\frtrain{} with different amounts of poisoning}
Table~\ref{tbl:poisoning} shows \frtrain{} performances with the different levels of poisoning. Even on the heavily poisoned (say 40\%) data, \frtrain{} shows marginal performance degradations ($<6.5\%$ decrease in DI).

\begin{table}[htpb]
  \caption{Accuracy and fairness performances of \frtrain{} on the poisoned synthetic test datasets for different amount of poisoning. We used the same label poisoning attack described in Section~\ref{sec:vulnerability}.}
  \label{tbl:poisoning}
  \centering
  \begin{tabular}{llcc}
    \toprule
    Data & Poisoning amount & DI & Accuracy  \\
    \midrule
    \multirow{1}{*}{Clean} & {\sc 0\%} & 0.818 & 0.807\\
     \cmidrule(r){1-4}
     \multirow{7}{*}{Poisoned} &
     {\sc 10\%}  & 0.827 & 0.814 \\
     &{\sc 15\%}  & 0.813 & 0.800 \\
     &{\sc 20\%}  & 0.802 & 0.800 \\
     &{\sc 25\%}  & 0.784 & 0.803 \\
     &{\sc 30\%}  & 0.780 & 0.800 \\
     &{\sc 35\%}  & 0.770 & 0.802 \\
     &{\sc 40\%}  & 0.765 & 0.806 \\
    \bottomrule
  \end{tabular}
\end{table}

\paragraph*{Meta learning with different validation set sizes}
Table~\ref{tbl:meta learning} shows the accuracy and fairness results for RML for different validation set sizes. We observe a drastic decrease of accuracy and fairness when the validation set size is 0.1\% of the training data.

\begin{table}[htpb]
  \caption{Accuracy and fairness performances of the meta learning method by~\cite{DBLP:conf/icml/RenZYU18} on the clean and poisoned synthetic test datasets for different validation set sizes. We used the same label poisoning attack described in Section~\ref{sec:vulnerability}, and the amount of poisoning is 10\% of $\mathcal{D}_{tr}$.}
  \label{tbl:meta learning}
  \centering
  \begin{tabular}{llcc}
    \toprule
    Data & Val. set size &  Disparate impact & Accuracy  \\
    \midrule
     \multirow{1}{*}{Clean} & {\sc 10\%} & 0.429 & 0.883\\
     \cmidrule(r){1-4}
     \multirow{5}{*}{Poisoned} &
     {\sc 10\%}  & 0.395 & 0.869 \\
     &{\sc 5\%}  & 0.378 & 0.852 \\
     &{\sc 0.5\%}  & 0.290 & 0.830 \\
     &{\sc 0.1\%}  & 0.098 & 0.714 \\
    \bottomrule
  \end{tabular}
\end{table}

\subsubsection{Real data}

We continue our experiments from Sections~\ref{sec:vulnerability} and~\ref{sec:realdata}.

\begin{figure}[t]
\centering
\begin{subfigure}{\columnwidth}
\centering
\includegraphics[scale=0.33]{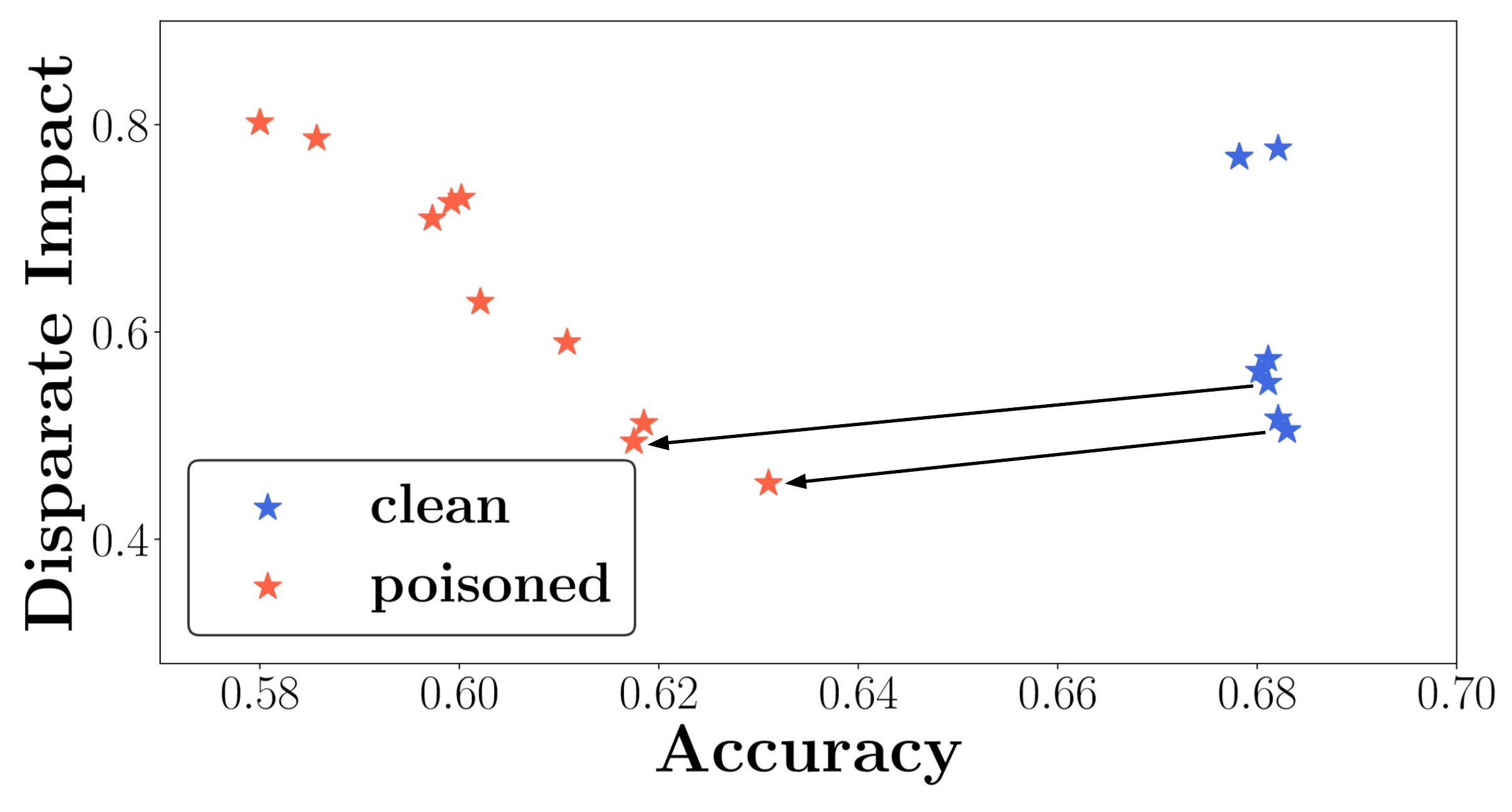}
\caption{Accuracy-fairness tradeoff curve on COMPAS dataset}
\label{fig:fc_compas}
\end{subfigure}
\begin{subfigure}{\columnwidth}
\centering
\includegraphics[scale=0.33]{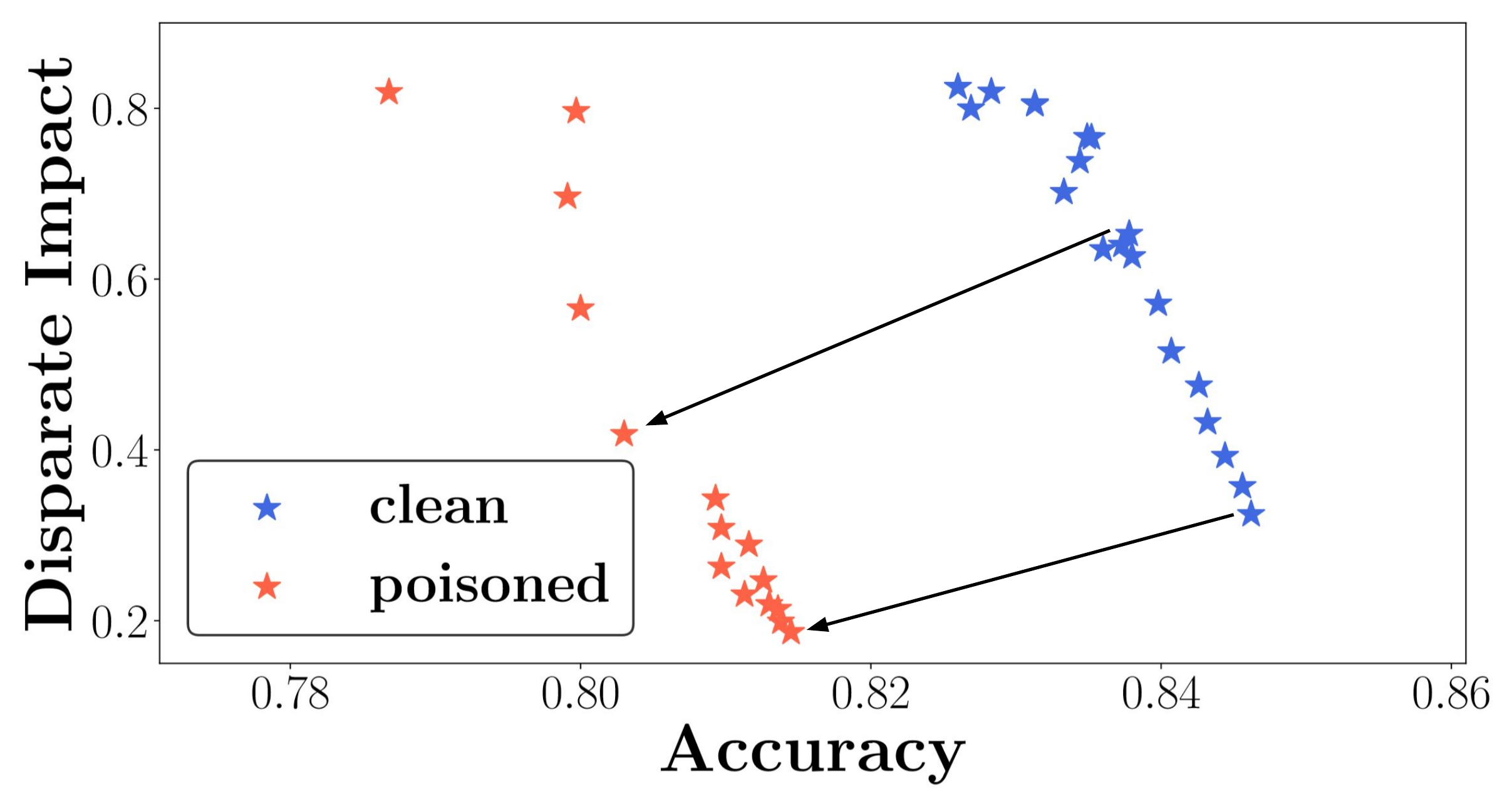}
\caption{Accuracy-fairness tradeoff curve on AdultCensus dataset}
\label{fig:fc_adultcensus}
\end{subfigure}
\caption{Accuracy-fairness tradeoff curves of Fairness Constraints on real datasets.}
\vspace{-0.5cm}
\end{figure}

\paragraph*{Fairness Constraints on real datasets}
We show the accuracy-fairness tradeoffs of Fairness Constraints~\cite{DBLP:conf/aistats/ZafarVGG17} on the COMPAS and AdultCensus datasets. Figures~\ref{fig:fc_compas} and~\ref{fig:fc_adultcensus} show that both accuracy and fairness of Fairness Constraints decrease on the poisoned data, showing a strictly-worse tradeoff.

\paragraph*{Training with only validation set}
We evaluate the baseline that simply trains fairness algorithms on the clean validation set. Table~\ref{tbl:validation} shows that the baseline performs worse than those in Tables~\ref{tbl:realCompas} and~\ref{tbl:realAdult}. For example, training FC on the AdultCensus crowdsourced validation set yields (DI, Acc) = (0.756, 0.761), which is worse than the FC baseline result (DI, Acc) = (0.826, 0.825) as shown in Table~\ref{tbl:realAdult}. We thus observe that the validation set is sufficient to help discern clean and poisoned data in \frtrain{}, but not large enough for algorithms to obtain high performance.

\begin{table}[t]
  \caption{Accuracy and fairness performances of the baseline that trains with only validation set. We use the same validation sets utilized in \frtrain{}.}
  \label{tbl:validation}
  \centering
  \begin{tabular}{lcccc}
    \toprule
     Method & \multicolumn{2}{c}{COMPAS} & \multicolumn{2}{c}{AdultCensus}\\
    \cmidrule(r){1-5}
      & DI & Acc. & DI & Acc. \\
    \midrule
     FC & 0.796 & 0.647 & 0.761 & 0.756 \\
     LBC & 0.796 & 0.647 & 0.795 & 0.799 \\
     AD & 0.762 & 0.646 & 0.682 & 0.693 \\
    \bottomrule
  \end{tabular}
  \vspace{-0.1cm}
\end{table}

\paragraph*{\frtrain{} using other fairness measures}
As we showed in Appendix~\ref{sec:eo_extension}, \frtrain{} respects equalized odds and equal opportunity. Table \ref{tbl:equalized odds} shows the experimental results on the synthetic and real datasets for equalized odds. We see that \frtrain{} significantly improves equalized odds with reasonable accuracy. The results w.r.t.\@ equal opportunity are similar and thus not shown here.

\begin{table}[t]
  \caption{Accuracy and fairness performances on synthetic and real test datasets w.r.t.\@ equalized odds. Two algorithms are compared: (1) LR (non-fairness method) and (2) \frtrain{}.}
  \label{tbl:equalized odds}
  \centering
  \begin{tabular}{llccc}
    \toprule
    Dataset & Method & \multicolumn{2}{c}{Equalized odds} & Accuracy  \\
    \cmidrule(r){1-5}
      &  & $Y$ = 0 & $Y$ = 1 &   \\
    \midrule
    \multirow{2}{*}{Synthetic Data} & LR  & 0.351 & 0.804 & 0.885 \\
     & \frtrain{} & 0.888 & 0.936 & 0.865\\
    \cmidrule(r){1-5}
    \multirow{2}{*}{COMPAS} & LR  & 0.427 & 0.557 & 0.674 \\
     & \frtrain{} & 0.718 & 0.959 & 0.628\\
    \cmidrule(r){1-5}
    \multirow{2}{*}{AdultCensus} & LR  & 0.286 & 0.909 & 0.848 \\
     & \frtrain{} & 0.503 & 0.917 & 0.842\\
    \bottomrule
  \end{tabular}
  \vspace{-0.1cm}
\end{table}

\subsection{Training methodology}
\label{sec:trainingmethodology}

The generator $G$ is a neural network with zero or one hidden layer. The discriminator $D^f$ is a single layer neural network, and the discriminator $D^r$ is a neural network with one hidden layer. We used 8 or 16 nodes in the hidden layers. We set an Adam optimizer~\cite{kingma2014adam} for the generator, and a stochastic gradient descent (SGD) optimizer for each discriminator. 
We empirically observe that one can stabilize the training procedure by freezing the parameters of the fairness discriminator $D^f$ for the initial phase of training.
Thus, we choose to freeze the parameters of the fairness discriminator $D^f$ for the first few epochs until the generator achieves a certain accuracy.
We pre-train the generator for the first few epochs and use the generator/discriminator update ratio of 1:3 (or 1:5) for the rest of training.

Also, we use the following details for choosing the values of $\lambda_1$, $\lambda_2$, and $C$. 
For clean data, we set $\lambda_2$ as a small value (e.g., $0.1$) and vary $\lambda_1$ from $0$ to $0.85$. 
For poisoned data, we set $\lambda_2$ as $0.2$, $0.3$, or $0.4$, and vary $\lambda_1$ from $0$ to $0.95-\lambda_2$. 
Given the values of $\lambda_1$ and $\lambda_2$, we also normalize $L_1$ (the generator loss) by multiplying it with $(1-\lambda_1-\lambda_2)$.
We set $C$ to be a value between $0$ to $3$.

\end{document}